\documentclass[times, review, 10pt]{elsarticle}
\usepackage[numbers]{natbib}
\usepackage{hyperref}
\hypersetup{hypertex=true,
            colorlinks=true,
            linkcolor=blue,
            anchorcolor=blue,
            citecolor=blue}

\usepackage{graphicx}
\usepackage{CJKutf8}
\usepackage{float}
\usepackage{multirow}
\usepackage{enumitem}
\usepackage{algorithm}
\usepackage{algpseudocode}
\newcommand{\myref}[1]{%
  \hyperref[#1]{Eq.~(\ref*{#1})}%
}

\usepackage{amssymb}
\usepackage{amsmath}
\journal{arXiv}

\begin{document}

\begin{frontmatter}


\title{UR2P-Dehaze: Learning a Simple Image Dehaze Enhancer via Unpaired Rich Physical Prior}

\author[1]{Minglong Xue} %% Author name
\author[1]{Shuaibin Fan}
\author[2]{Shivakumara Palaiahnakote}
\author[3]{Mingliang Zhou}
\affiliation[1]{organization={College of Computer Science and Engineering},%Department and Organization
            addressline={Chongqing University of Technology}, 
            city={Chongqing University of Technology},
            %postcode={400054}, 
            country={China}}
\affiliation[2]{organization={School of Science, Engineering and Environment},%Department and Organization
            addressline={ University of Salford}, 
            city={Manchester},
            % postcode={400044}, 
            country={UK}}
\affiliation[3]{organization={College of Computer Science},%Department and Organization
            addressline={Chongqing University}, 
            city={Chongqing},
            %postcode={400044}, 
            country={China}}
%% Abstract
\begin{abstract}
Image dehazing techniques aim to enhance contrast and restore details, which are essential for preserving visual information and improving image processing accuracy. Existing methods rely on a single manual prior, which cannot effectively reveal image details. To overcome this limitation, we propose an unpaired image dehazing network, called the Simple Image Dehaze Enhancer via Unpaired Rich Physical Prior (UR2P-Dehaze). First, to accurately estimate the illumination, reflectance, and color information of the hazy image, we design a shared prior estimator (SPE) that is iteratively trained to ensure the consistency of illumination and reflectance, generating clear, high-quality images. Additionally, a self-monitoring mechanism is introduced to eliminate undesirable features, providing reliable priors for image reconstruction. Next, we propose Dynamic Wavelet Separable Convolution (DWSC), which effectively integrates key features across both low and high frequencies, significantly enhancing the preservation of image details and ensuring global consistency. Finally, to effectively restore the color information of the image, we propose an Adaptive Color Corrector that addresses the problem of unclear colors. The PSNR, SSIM, LPIPS, FID and CIEDE2000 metrics on the benchmark dataset show that our method achieves state-of-the-art performance. It also contributes to the performance improvement of downstream tasks. The project code will be available at \url{https://github.com/Fan-pixel/UR2P-Dehaze}.
\end{abstract}
%\href{https://github.com/Fan-pixel/UR2P-Dehaze}{Link}

% \begin{highlights}
% \item We propose a Shared Prior Estimator that adaptively learns rich physical priors during the dehazing process.
% \item We design a Dynamic Wavelet Separable Convolution to enhance multi-scale feature extraction capabilities.
% \item We introduce an Adaptive Color Corrector to iteratively learn RGB channel information.
% \end{highlights}

%% Keywords
\end{frontmatter}




\section{Introduction}
\label{sec1}
%% Labels are used to cross-reference an item using \ref command.

With the rapid development of computer vision technology, image dehazing has become an important research focus. This technique is crucial in fields like object detection and traffic monitoring, where weather conditions such as haze can significantly reduce image clarity. Therefore, improving image quality by removing haze and restoring details has become particularly urgent. Additionally, enhancing image contrast not only benefits human visual perception but also effectively improves the performance of machine vision systems.

In recent years, deep learning technology has made significant progress in the field of image dehazing, improving the performance of advanced visual tasks such as object detection (such as RTDET\cite{zhao2024detrs}, Far3D\cite{jiang2024far3d}, DR-YOLO\cite{zhong2024dehazing}), recognition and semantic segmentation, and showing advantages in lower-level visual tasks such as image super-resolution (such as Epico\cite{xie2023sepico}, IDM\cite{gao2023implicit}), de-noise (such as MWDCNN\cite{tian2023multi}, LPDM\cite{panagiotou2024denoising}, RDDM\cite{liu2024residual}) and enhancement (such as GSAD\cite{hou2024global}, SEMACOL\cite{niu2025semacol}). To better understand the formation of haze in images, the atmospheric scattering model is often used as a theoretical foundation. Nevertheless, image dehazing methods based on deep learning still need to be improved when dealing with complex scenes, especially in terms of strengthening model generalization ability. The atmospheric scattering model equations can be written as~\myref{eq11}:
\begin{equation}\label{eq11}
I(x)=J(x)t(x)+A(x)(1-t(x)),
\end{equation}
Where $ I(x) $ and $ J(x) $ represent the observed hazy image and clean image, $ t(x) $ and $ A(x) $ are the global atmospheric light and transmission map, respectively. Usually defining $t(x)=\boldsymbol e^{-\beta d(x)}$. The $ \beta $ represents the atmospheric scattering coefficient, and $ d(x) $ is the scene depth. \par 
Existing dehaze methods mainly rely on hand-designed prior knowledge, such as atmospheric light and depth information, and these methods have limitations in practical applications. First of all, due to the complexity of haze weather, single manual prior knowledge is difficult to adapt to different haze conditions, resulting in an unsatisfactory dehazing effect. Secondly, these methods usually require multiple input images and use the information differences between image to recover image details, but it is often difficult to obtain paired hazy and haze-free images in practical applications. In addition, the amount of information in a single image is limited, and manual prior knowledge cannot make full use of this information to recover image details.
\begin{figure*}
\setlength{\abovecaptionskip}{-0.6cm}   %调整图片标题与图距离
\centering 
\includegraphics[width=\textwidth]{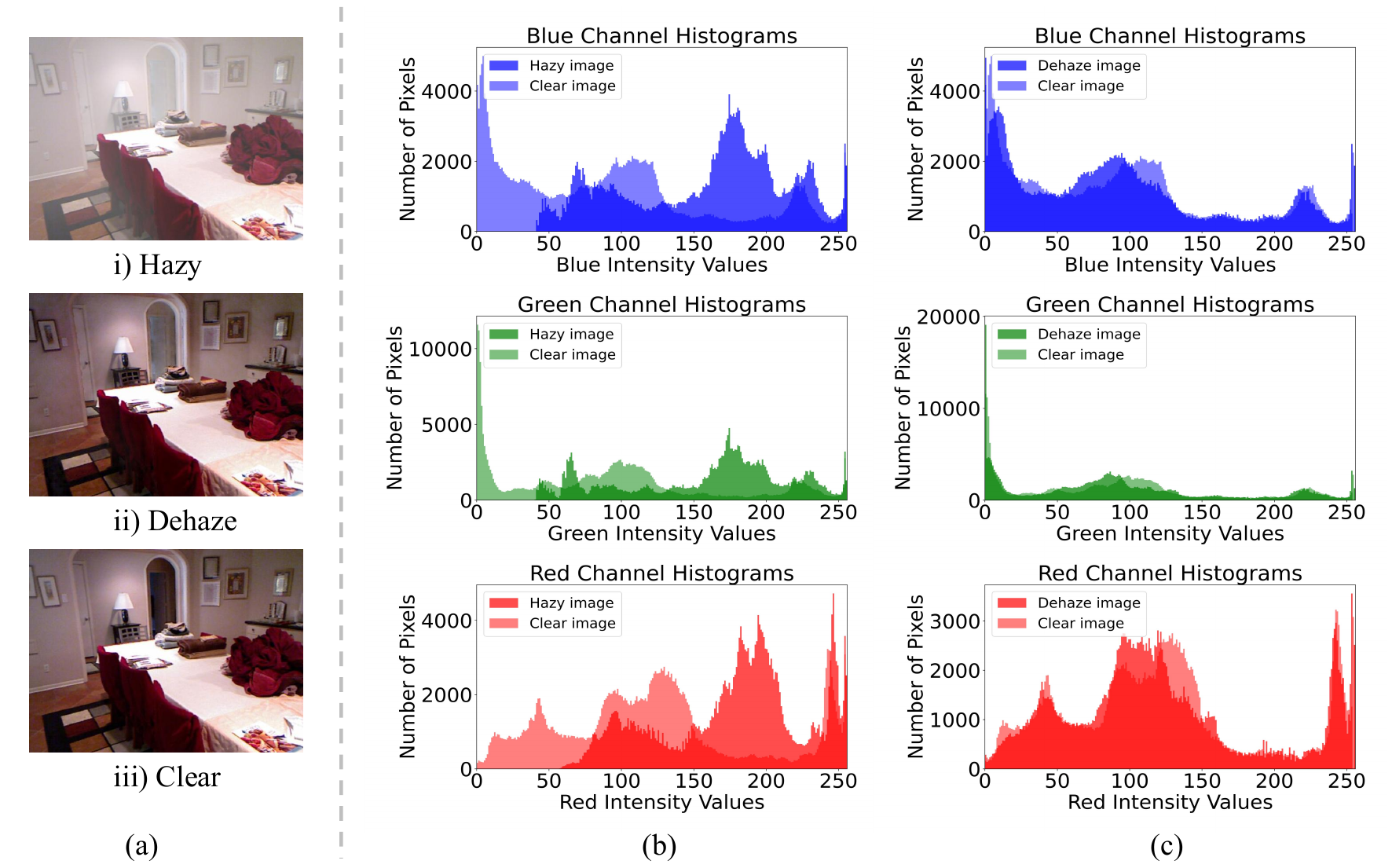} % 插入图片，假设图片文件名为 example-image.png
\caption{RGB three-channel comparison of the image with haze and the image with the haze removed. (a) Represents the visual comparison between the hazy image, the dehaze image, and the clear image. (b) Representation of the difference in the RGB three channels between the hazy image and the clear image. (c) Difference between the dehaze image and the clear image in the RGB three channels.} 
\label{fig:fig1} 
\end{figure*}

In recent years, although many researchers have conducted extensive studies on dehazing using deep learning methods, our extensive experiments reveal that few studies focus on color restoration. However, color restoration plays a crucial role in image dehazing tasks. In order to verify this view, we compared the differences between the hazy image and the original dehazed image on the three RGB channels in~\autoref{fig:fig1} (b). Similarly, in~\autoref{fig:fig1} (c), we compare the performance of the dehazed image generated by D4~\cite{yang2022self} with the original dehazed image on the RGB channels. The experimental results show that the values of the dehazed image generated by D4~\cite{yang2022self} on the three RGB channels are not consistent with the values of the original clear image, and the color and luminance distributions are also different.\par
\begin{figure*}[h]
\setlength{\abovecaptionskip}{-0.6cm}   %调整图片标题与图距离
\centering 
\includegraphics[width=\textwidth]{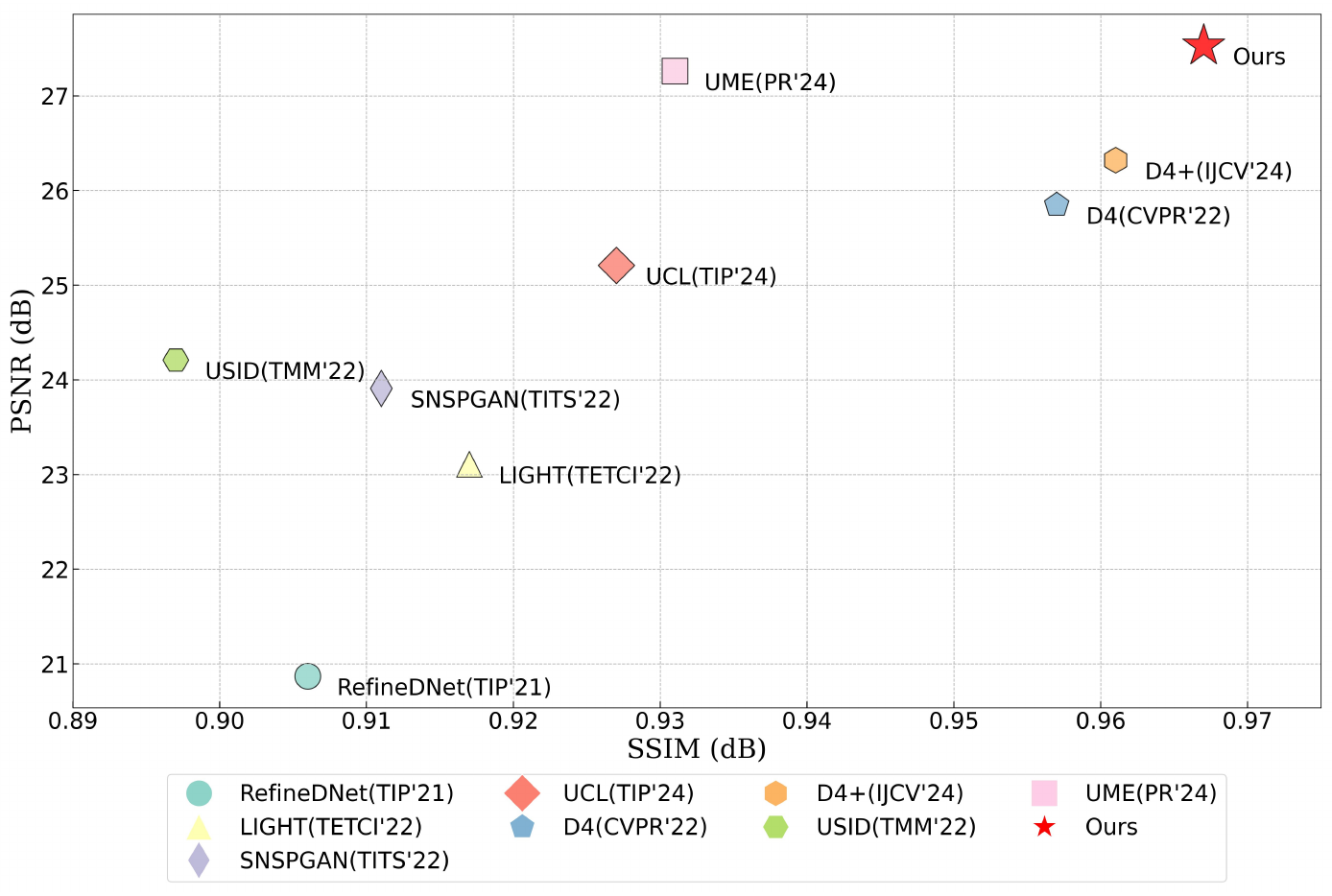} % 插入图片，假设图片文件名为 example-image.png
\caption{The average PSNR and SSIM are used as evaluation metrics to compare the performance of the proposed unsupervised dehazing method with other state-of-the-art methods on the SOTS-outdoor dataset.} 
\label{fig:PSNRSSIM} 
\end{figure*}
To address the aforementioned issues, we propose an unsupervised adaptive learning-based dehazing method called UR2P-Dehaze, which is grounded in rich prior information. The core idea of UR2P-Dehaze is to learn rich prior estimation from hazy images, including color estimation, reflectance estimation, and illumination estimation. Additionally, UR2P-Dehaze utilizes a perfect combination of convolutions and depthwise separable convolutions in the wavelet domain during the dehazing process to obtain a larger receptive field, thus more effectively capturing both local features and global information in the image. As an effective multi-resolution analysis tool, wavelet transform captures multi-scale features of the image through analysis at different scales, which is particularly crucial for image dehazing. Furthermore, the Adaptive Color Corrector is designed in the image reconstruction process to better learn color information. This approach allows UR2P-Dehaze to better adapt to different haze conditions, preserving more details during image processing and improving the accuracy and naturalness of haze removal. As shown in~\autoref{fig:PSNRSSIM}, our method achieves the best results compared to other state-of-the-art methods. These findings emphasize the importance of effective color estimation in image dehazing tasks, which will be further validated in the experimental section.\par
Moreover, UR2P-Dehaze features a concise network structure while efficiently learning rich prior knowledge, thereby enhancing the method's generalization ability and practical application value. Experimental results on public datasets (SOTS-indoor, SOTS-outdoor, I-HAZE, and HSTS) demonstrate that UR2P-Dehaze performs well, verifying its effectiveness and potential in the field of image dehazing. The main contributions of this work include:
\begin{itemize}[noitemsep,topsep=10pt]
\item We propose a novel unpaired image dehazing network, UR2P-Dehaze, which automatically learns rich physical prior, such as reflectance and illumination, from hazy image, significantly enhancing the adaptability and generalization ability of dehazing methods.  
\item UR2P-Dehaze employs convolution operations in the wavelet domain during the dehazing process, effectively expanding the network's receptive field and optimizing the capture of both local and global image features. This plays a crucial role in restoring image details and improving the dehazing effect.  
\item Through experimental validation of the model's sensitivity to color, we propose the Adaptive Color Corrector, which enables iterative learning of color information.  
\end{itemize}
%% Use \subsection commands to start a subsection.

The rest of this article is organized as follows: Section~\ref{sec:section2} provides an overview of related work on haze removal. Section~\ref{sec:section3} details the proposed methods, including SPE, DWSC, and ACC. Section~\ref{sec:section4} presents the results of a series of ablation experiments. Finally, Section~\ref{sec:section5} concludes the article.
\section{Related Work}
\label{sec:section2}
The research on image dehazing technology is mainly divided into five directions: traditional methods, semi-supervised deep learning methods, supervised deep learning methods, unsupervised deep learning methods, and Retinex theory-based dehazing methods. We will describe these three methods in detail below.
\subsection{Traditional Learning Dehazing Methods}
\label{subsec1}

The key to these image dehazing techniques based on traditional methods is to identify and calculate the key parameters in the imaging process. They utilize artificially constructed prior information such as dark channel prior (DCP) \cite{he2010single}, color decay prior (CAP) \cite{zhu2015fast}, and gamma correction prior (GCP) and integrate it into the atmospheric scattering model (ASM) \cite{mccartney1976optics} to infer image transmittance and ambient illumination accurately. This method can effectively restore clear image without haze.
In the early studies of image dehazing techniques, these methods mainly relied on physical models and image statistical properties to recover images. Specifically, the dark channel prior (DCP) theory proposed by He et al.\cite{he2010single} uses the information of dark channels in images to estimate atmospheric light, providing an effective statistical method for haze removal. These traditional methods have laid the foundation in the field of haze removal. Still, they are limited by the design of manual features and rely on simplified physical models, which are difficult to adapt to complex haze conditions.

%% Use \subsubsection, \paragraph, \subparagraph commands to 
%% start 3rd, 4th and 5th level sections.
%% Refer following link for more details.
%% https://en.wikibooks.org/wiki/LaTeX/Document_Structure#Sectioning_commands
\subsection{Supervised Learning Dehazing Methods}
With the rise of deep learning technology, the supervised deep learning method has made remarkable progress in the field of image dehazing\cite{zheng2023curricular,chen2024dea,cui2023strip}. These methods are trained on a large number of pairs of hazy and non-hazy images, learning complex mapping relationships from hazy images to clear images. The end-to-end network proposed by Berman et al. \cite{berman2016non} uses deep convolutional neural networks (CNNs) to learn dehaze directly from hazy images, which significantly improves the dehaze effect. Mishra et al. \cite{mishra2024h2cgan} used unsupervised generative adversarial networks to improve visual quality and target detection accuracy in hazy and rainy images, achieving significant performance improvements. Work shown by~\cite{li2017aod} proposed an end-to-end AODNet dehazing network that produces a clear image by reformulating the atmospheric scattering model. However, it requires a large number of image pairs with and without haze, which is often not feasible in practical applications.
\subsection{Semi-supervised Learning Dehazing Methods}
In the field of dehazing, semi-supervised learning methods\cite{jia2023semi} have attracted attention because of their effective use of limited labeled data and large amounts of unlabeled data. Among them, the SADNet method \cite{sun2022sadnet} is a typical example, which combines attention mechanisms to improve the effect of removing haze from a single image. SADNet\cite{sun2022sadnet} enables models to learn the ability to recover clear images from haze through supervised training on synthetic data and unsupervised training on real-world data. The advantage of this method is that it can not only learn specific features of dehazing from labeled data but also learn more generalized features from a large number of unlabeled data, thus improving the robustness and adaptability of dehazing methods.

Another semi-supervised dehazing method is implemented by building a multi-branch learning framework. The semi-supervised image dehazing network \cite{an2022semi} consists of supervised branches, the former using coding-decoding neural network structure and constrained by supervised loss, and the latter using prior knowledge to construct unsupervised loss to estimate transmission graphs and atmospheric light. The two branches output dehazing results and constrain the network by minimizing reconstruction losses, making it more generalized. However, when the prior knowledge is used to estimate parameters, the performance of the model is unstable because the processing of the prior information is not fine enough.
\subsection{Unsupervised Learning Dehazing Methods}

In recent years, unsupervised learning methods have made remarkable progress in the field of image dehazing\cite{li2022usid,sun2024unsupervised,ding2023u}, which solves the problem of dependence of traditional methods on paired clear and hazy image datasets. Engin et al. \cite{engin2018cycle} proposed a haze removal framework based on CycleGAN\cite{zhu2017unpaired}, which realized the conversion from a hazy image to a clear image through cyclic consistency antagonism training while maintaining semantic consistency. Shao et al. \cite{shao2020domain} combined domain adaptive technology to adjust the feature distribution of synthetic and real hazy image, thus enhancing the generalization ability of the model under diverse haze conditions. However, these methods are difficult to effectively recover image details due to limited information and strong dependence on artificial prior knowledge when processing single image, resulting in limited adaptability and generalization ability.

\subsection{Retinex Theoretical Learning Dehazing Methods}
In the field of image dehazing, Retinex theory has been widely used \cite{huang2025image,yi2023diff}. Applications of Retinex theory~\cite{fu2023learning} include single-scale Retinex method and multi-scale Retinex method. Among them, the single-scale Retinex method is more sensitive to the scale selection of the surrounding function, while the multi-scale Retinex method overcomes this problem by estimating the illumination composition at multiple scales. However, the main challenge with these methods is the need to manually select the appropriate surrounding functions and their corresponding scaling parameters. Gui et al.\cite{gui2023illumination} proposes IC-Dehazing, a multi-output dehazing network with illumination controllable ability, which allows users to adjust the illumination intensity for dehazed image and is based on the interpretable Retinex model. In addition, Li et al. \cite{9274531} proposed a deep retinex dehazing network (RDN) based on Retinex theory. The multi-scale residual network was used to estimate the residual light pattern and combined with U-Net with channel and spatial attention mechanism to achieve high-precision restoration of haze-free image. To address these challenges, we propose a novel network architecture that overcomes scale constraints and enables the learning of richer and more precise estimation information.
\begin{figure*}[h]
\setlength{\abovecaptionskip}{-0.6cm}   %调整图片标题与图距离
\centering 
\includegraphics[width=\textwidth]{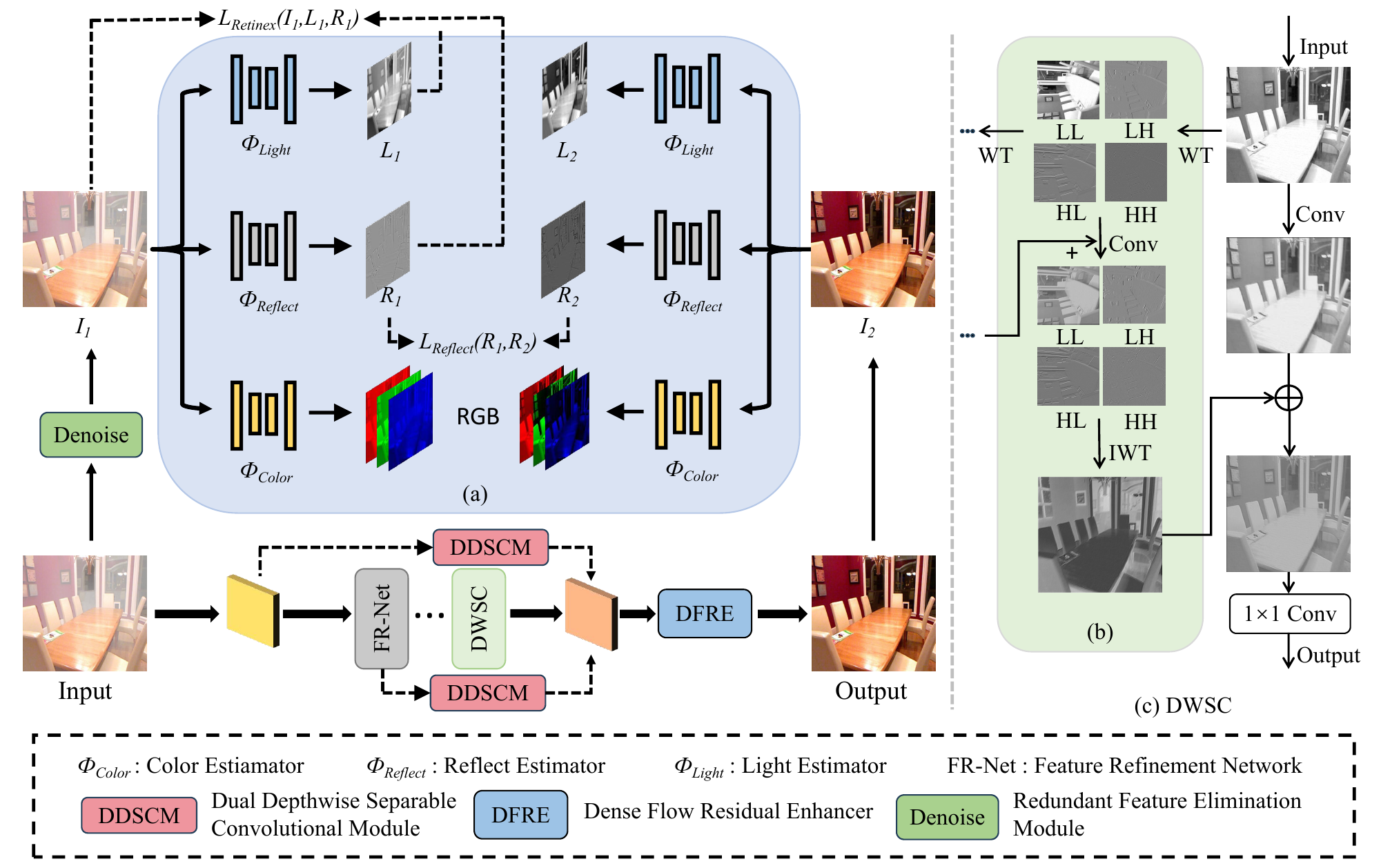} % 插入图片，假设图片文件名为 example-image.png
\caption{The overall pipeline of our UR2P-Dehaze. It comprises (a) an adaptive learning branch that estimates illumination, reflectance, and color priors, (b) a cascaded frequency decomposition process using wavelet transform, and (c) The dynamic wavelet separable convolution (DWSC) module leverages physical features to dynamically model the image, adaptively combining frequency and spatial domain information to refine dehazing results and enhance realism and color fidelity. Additionally, the feature refinement network incorporates both a residual module and a feature fusion module to further enhance feature representation and integration.} 
\label{fig:overall_frame} 
\end{figure*}

\section{Mathematics}
\label{sec:section3}
In this section, we first introduce the Shared Prior Estimator (SPE) in Sections~\ref{SPE} and ~\ref{ACC} to guide the model in learning key image parameters during the dehazing process, thereby enhancing the model's ability to accurately capture contrastive parameters such as illumination, reflectance, and color information. These learned parameters provide reliable prior support for the subsequent image reconstruction. In Section~\ref{DWSC}, during the image reconstruction stage, we further incorporate Dynamic Wavelet Separable Convolution (DWSC), which performs convolution operations across different frequency components and employs a dynamic scaling mechanism. This effectively integrates the previously learned illumination, reflectance, and other key information, significantly improving the detail preservation and global consistency of the reconstructed images. The overall framework of our method is shown in~\autoref{fig:overall_frame}.
\subsection{Shared Prior Estimator}
\label{SPE}
\textbf{Prior Estimation Guided by Retinex Theory.} The Retinex theory\cite{land1977retinex} was proposed in the early years to explain how the human visual system perceives color and brightness. The central idea of this theory is that the color of an object is determined by its ability to reflect light of different wavelengths, rather than the absolute intensity of the reflected light. It is based on two main concepts: the three-color theory and color constancy. According to the three-color theory, the color of an object is determined by its ability to reflect red, green, and blue light, while color constancy means that the color of an object is not affected by uneven lighting and can remain consistent even under different lighting conditions. The Retinex theory can be expressed as $I = L\circ R$.
% \begin{equation}
% I = R * L,
% \label{eq2}
% \end{equation}
The Retinex theory further developed a variety of image enhancement methods, including single-scale Retinex (SSR) and multi-scale Retinex (MSR) with color recovery (MSRCR). By simulating the human eye's adaptability to light changes, these methods aim to improve the visual quality of images under different lighting conditions, enhancing the contrast and brightness of images, while retaining detailed information. Retinex algorithm has a wide range of applications in image processing, such as illumination restoration, color constancy, image enhancement, and medical image processing.

However, due to the diversity of natural scenes and lighting conditions, hand-crafted priors are often not sufficient to accommodate. Moreover, the illumination information in the recovery results is particularly uneven. In this paper, instead of utilizing manual priors for $L$ and $R$ from a single image, we apply pairs of hazy images to automatically learn adaptive priors in a data-driven manner. These hazy images share the same scene content but have different lighting conditions. Mathematically, the Retinex decomposition of hazy image pairs can be expressed as:
\begin{equation}
I_{hazy}=L_{hazy} \circ R,
\end{equation}
Where the hazy image \( I_{hazy} \) can be decomposed into illumination \( L_{hazy} \) and reflectance \( R \), with \( \circ \) representing the element-wise product.

In the haze removal task, the traditional Retinex theory relies on hand-crafted prior knowledge to estimate the illumination and reflection components in the image. However, due to the diversity of natural scenes and lighting conditions, this manual prior approach is often difficult to adapt to all situations. To overcome this limitation, this paper proposes a data-driven approach that automatically iteratively learns adaptive priors from pairs of haze and recovered images. These hazy image pairs share the same scene content but have different lighting conditions, allowing the network to learn more estimated information without being constrained by a specific size. Mathematically, the Retinex decomposition of a pair of images with haze can be expressed as~\myref{eq4}:
\begin{equation}
\left\{
\begin{array}
{c}I_{hazy}{=}L_{hazy}\circ R \\
I_{clear}{=}L_{clear}\circ R
\end{array}\right.,
\label{eq4}
\end{equation}
Where $ I_{hazy} $ and $ I_{clear}$ represent the hazy image and recovery maps in the iterative process, while $ L_{hazy} $ and $ L_{clear} $ denote the illumination image during the iteration. They share the same reflectance component $ R $.

\textbf{Projection Loss.} Before hazy image are input into the network, it is essential to preprocess the image to eliminate ill-posed characteristics caused by atmospheric scattering and lighting conditions. These characteristics typically include uneven illumination, low contrast, and color distortion, which can negatively impact the performance of subsequent image enhancement and decomposition methods. Effective preprocessing can significantly improve the Retinex model's ability to decompose the image, allowing it to more accurately separate the reflectance and illumination components.

\begin{equation}
\mathcal{L}_{project}=\left\|I_{hazy}-I_{project}\right\|_2^2,
\end{equation}
Where \( I_{project} \) denotes the image after removing redundant features.

Additionally, some irrelevant information will be removed, which can be explained using the following formula.
\begin{equation}
\begin{aligned}
& \operatorname*{\mathrm{argmin}}_{L,R}\|I_{hazy}-L\circ R-\delta\| =\underset{L,R}{\operatorname*{\operatorname*{argmin}}}\|I_{hazy}-I_{project}+I_{project}-L\circ R-\delta\|     \\
& \hspace{4.3cm}\leq\|I-I_{project}-\delta\|+\underset{L,R}{\operatorname*{\operatorname*{argmin}}}\|I_{project}-L\circ R\|.
\end{aligned}
\end{equation}
Where $ \delta $ represents the error, it is important to note that projection loss must be combined with other constraints to avoid trivial solutions, such as \( I_{hazy} = I_{project} \).

\textbf{Reflectance Consistency Loss.} In the field of dehazing, we calculate a new reflection consistency loss ($\mathcal{L}_{reflect}$) based on paired low-light image pairs and Retinex theory. Compared to prior knowledge constructed by hand, $\mathcal{L}_{reflect}$ is more accurate and adaptable because it is based on the physical properties of the object. Mathematically, $\mathcal{L}_{reflect}$ is defined as~\myref{eq7}:
\begin{equation}
\mathcal{L}_{reflect}=\left\|R_1-R_2\right\|_2^2,
\label{eq7}
\end{equation}
where $ R_1 $ and $ R_2 $ denote the reflectance maps predicted by the $\mathit{\Phi_{Reflect}}$ estimator and the reflectance maps of the rough dehazed image, respectively.

\textbf{Retinex Loss.} Retinex loss aims to measure the quality of the decomposition of the reflectance and illumination components, encouraging the network to learn a reasonable decomposition. Typically, this loss function optimizes the model by minimizing the difference between the predicted reflectance and illumination image and the actual image.
\begin{equation}
\begin{aligned}
\mathcal{L}_{retinex} & =\left\|R\circ L-I_{project}\right\|_{2}^{2}+\left\|R-I_{project}/stopgrad(L)\right\|_{2}^{2} \\
 & +\left\|L-L_{initial}\right\|_{2}^{2}+\left\|\nabla L\right\|_{1},
\end{aligned}
\end{equation}
Where $ L_{initial } $ represents the initially estimated illumination information, i.e. the maximum value for each RGB channel. To ensure that the decomposition of $ I_{project} $ is more accurate, $ \left\|R-I_{project}/stopgrad(L)\right\|_{2}^{2} $ is introduced to guide the decomposition.
\subsection{Dynamic Wavelet Separable Convolution}
\label{DWSC}
To expand the learning capacity of the dehazing enhancer while avoiding the exponential growth of parameters and increased model complexity caused by large convolution kernels, we propose Dynamic Wavelet Separable Convolution (DWSC). This method, based on wavelet transform theory, performs multi-scale feature extraction to emulate the effect of large-kernel convolutions. It combines the efficiency of depthwise separable convolutions with the frequency analysis capabilities of the wavelet transform, significantly improving performance while effectively controlling model parameterization.
\begin{equation}
I^{(i)}_{output}=IWT(Conv(W,WT(I^{(i)}_{input}))),
\end{equation}
Where $ I^{(i)}_{input} $ represents the input feature map at the $i-th$ layer, and $ I^{(i)}_{output} $ represents the feature map after convolution and inverse wavelet transform at the $i-th$ layer.
\begin{equation}
X_{LL}^{(i)},X_{H}^{(i)}=WT(X_{LL}^{(i-1)}),
\end{equation}
\begin{equation}
Y_{LL}^{(i)},Y_{H}^{(i)}=Conv(X_{LL}^{(i)},X_{H}^{(i)}),
\end{equation}
Where $ X_{H}^{(i)} $ and $ X_{LL}^{(i)} $ represent the return values of the wavelet transform applied to the low-frequency features from the previous layer. Specifically, $ X_{H}^{(i)} $ represents all the high-frequency features of the current layer, while $ X_{LL}^{(i)} $ represents the low-frequency features of the current layer. $ Y_{LL}^{(i)} $ and $ Y_H^{(i)} $ represent the features after convolution.
\begin{algorithm}[ht]
\label{algorithm1}
\caption{Dynamic Wavelet Separable Convolution}
\begin{algorithmic}[1]
\State \textbf{Input:}
\State \quad Hazy image tensor $X$ of shape $(B, C, H, W)$
\State \quad Wavelet filter type $wt\_type$ and levels $wt\_levels$
\State \textbf{Output:}
\State \quad Processed image $Y$
\State \textbf{Steps:}
\State 1. Initialize wavelet filters $(dec\_hi, dec\_lo, rec\_hi, rec\_lo)$ based on $wt\_type$.
\State 2. \textbf{For each level in $wt\_levels$:}
\State \quad\quad a. Apply wavelet transform to input $X$ to obtain low-frequency $(LL)$ and high-frequency components $(LH, HL, HH)$.
\State \quad\quad b. Perform depthwise separable convolution on $LL$ component.
\State \quad\quad c. Apply pointwise convolution on all components $(LL, LH, HL,$ $ HH)$.
\State \quad\quad d. Use inverse wavelet transform to reconstruct the output.
\State 3. Apply $base\_conv$ and scale output using a learnable weight scaling mechanism.
\State 4. Return the processed output $Y$.
% \State \textbf{Pseudocode:}
% \State \quad \texttt{Function WTConv2d(X):}
% \State \quad \quad Initialize wavelet filters
% \State \quad \quad \textbf{For each level in $wt\_levels$:}
% \State \quad \quad \quad Transform $X$ using wavelet filters
% \State \quad \quad \quad Process low-frequency $(LL)$ and high-frequency $(LH, HL, HH)$
% \State \quad \quad \quad Reconstruct output using inverse wavelet transform
% \State \quad \quad Apply base convolution and scale
% \State \quad \quad Return $Y$
% \State \texttt{Main:}
% \State \quad Initialize input $X$
% \State \quad Apply $WTConv2d$ to $X$
% \State \quad Output processed tensor $Y$
\end{algorithmic}
\end{algorithm}
\begin{equation}
Q^{(i)}=IWT(Y_{LL}^{(i)}+Q^{(i+1)},Y_{H}^{(i)}),
\end{equation}
Where $ Q^{(i)} $ represents the aggregated output result from the current layer to the next layer.

Specifically, DWSC is described in Algorithm 1, the module first uses wavelet decomposition to separate the input feature map into high-frequency and low-frequency components. It then applies small-kernel convolutions to these frequency components and fuses the processed features back into the original space through wavelet reconstruction. Additionally, the convolutional module includes a scaling component that uses learnable weight parameters to dynamically scale the feature map after wavelet transformation, enhancing the model’s adaptability to different frequency components and improving its feature representation capabilities.
\begin{figure*}[h]
\setlength{\abovecaptionskip}{-0.6cm}   %调整图片标题与图距离
\centering 
\includegraphics[width=\textwidth]{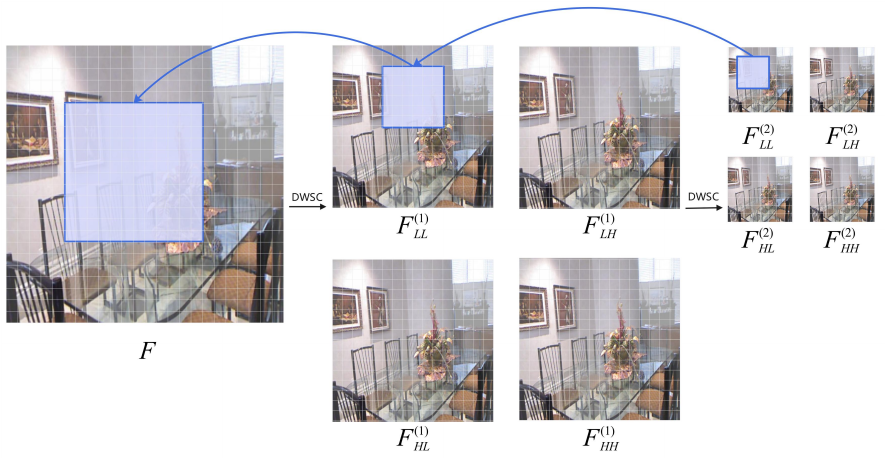} % 插入图片，假设图片文件名为 example-image.png
\caption{The process oof performing convolution operation in wavelet domain. It can effectively reduce the number of model parameters and improve the multi-scale feature extraction ability of the model.} 
\label{fig:DWSC} 
\end{figure*}
To recover frequency information at different depths and optimize both computational efficiency and parameter control, we combine depth-separable convolution with wavelet transforms to create the DWSC structure. In this structure, the traditional $3 \times 3$ convolution kernel is replaced by wavelet transform convolution, while pointwise convolution ($1 \times 1$ convolution) is used to integrate channel information. This design allows deep convolution to process each input channel independently in the wavelet domain, effectively capturing local features, while pointwise convolution fuses these features into the final output. As shown in~\autoref{fig:DWSC}, this approach not only reduces the number of parameters and computational load but also improves the network’s ability to capture local features in the wavelet domain, expanding the receptive field and enhancing feature representation, all while maintaining computational efficiency.

\subsection{Adaptive Color Corrector}
\label{ACC}
We aim to learn the color information that reveals the scene under varying haze thicknesses and use the color map $C_F$, predicted from the hazy feature image $F$, as pseudo-ground truth. We then train the color estimation network $\mathit{\Phi_{Color}}$ to estimate the color map $\hat{C}_F$ from the dehazed feature image $F_1$. Finally, we define the training losses:
\begin{equation}
\mathcal{L}_{color}=\left\|C_F-\hat{C}_F\right\|_1
\end{equation}
Where $C_F$ represents the RGB three-channel color estimation of the hazy image, and $\hat{C}_F$ represents the RGB three-channel color estimation of the haze-removed image during the training process.

\begin{figure*}
\setlength{\abovecaptionskip}{-0.6cm}   %调整图片标题与图距离
\centering
\includegraphics[width=\textwidth]{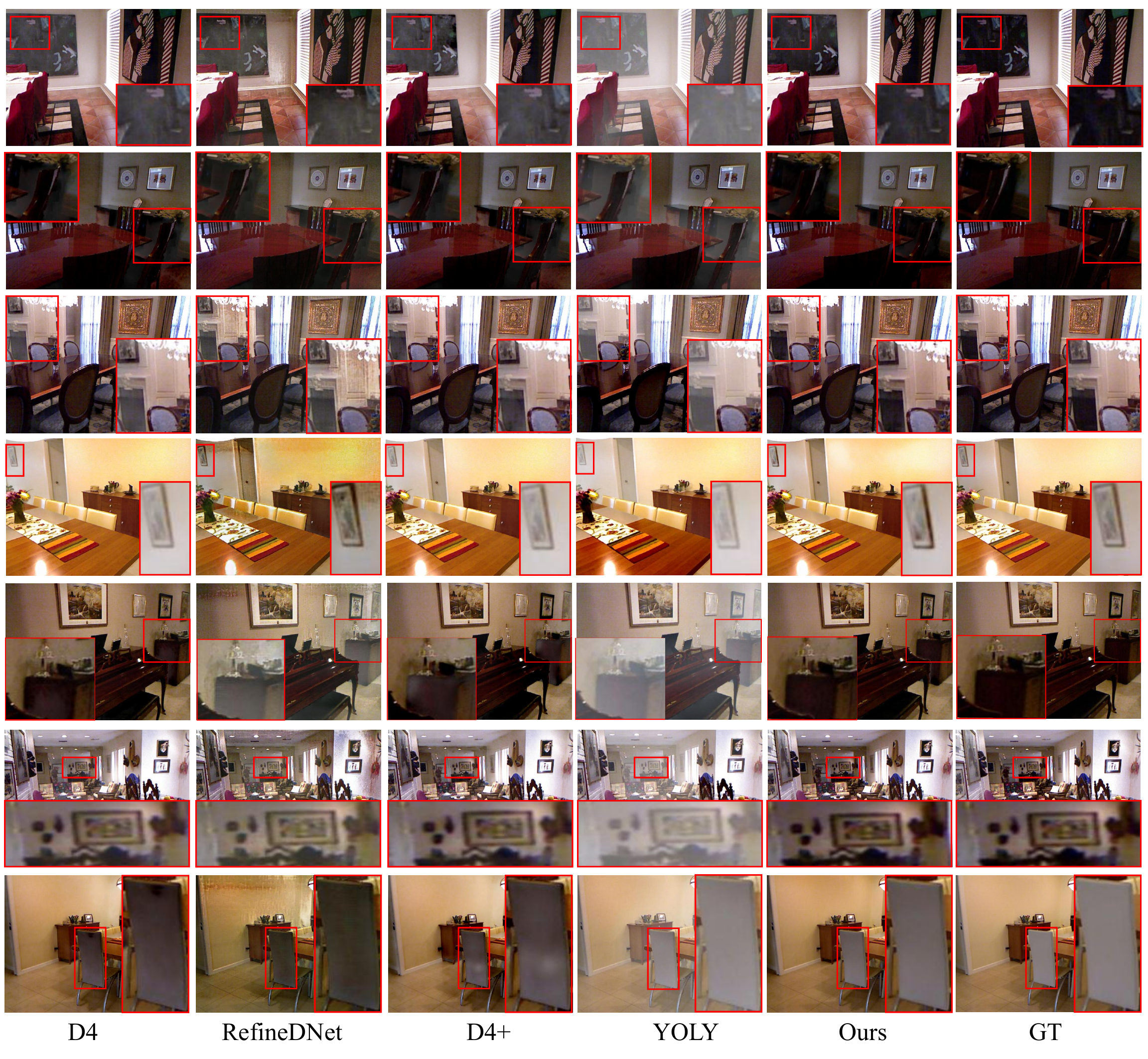}
\caption{Visual comparison of haze removal on samples from the SOTS-indoor dataset.}
\label{fig:indoor}
\end{figure*}

As shown in~\autoref{fig:indoor}, we can find that our proposed method is significantly better than other methods in terms of image color restoration.  Specifically, in the three RGB color channels, the distribution of pixel values in the restored image is closer to that of the clear image. This indicates that our method can restore image colors more accurately. In contrast, other methods exhibit biases in the color restoration process. For instance, the pixel values in the three channels of the image restored by the D4~\cite{yang2022self} method are generally higher, which may be due to the haze appearing whitish, resulting in color deviation during the restoration process. Furthermore, from another perspective, these methods still retain a certain degree of residual haze after dehazing, further highlighting their shortcomings in color restoration. Overall, our method demonstrates significant advantages in image color restoration, effectively removing haze and restoring the true colors of the image.

Further analysis of the results of the comparison between our method and other methods shows that our method performs better on the RGB three channels with the lowest difference value, which indicates that our method can recover the true color of the image more effectively. This result not only validates the importance of color recovery in dehazing tasks, but also highlights the advantages of our method in maintaining color authenticity and improving image quality. Overall, our method outperforms other SOTA methods in terms of color reproduction, which is important for improving the visual quality and usefulness of decolorized image.
\subsection{Overall Loss}
To sum up, our UR2P-Dehaze haze removal network uses $\mathcal{L}_{project}$, 
 $\mathcal{L}_{reflect}$, $\mathcal{L}_{retinex}$ and $\mathcal{L}_{color}$ loss functions, and the total loss function can be expressed as:
\begin{equation}
\mathcal{L}_{loss}=\lambda_1 \mathcal{L}_{project}+\lambda_2 \mathcal{L}_{reflect}+\lambda_3 \mathcal{L}_{retinex}+\lambda_4 \mathcal{L}_{color}.
\end{equation}
Where $\lambda_i$ represents different loss weights. In our experiments, we set $\lambda_1$ = 50, $\lambda_2$ = 0.1, $\lambda_3$ = 0.1 and $\lambda_4$ = 1 according to the experimental settings.
\begin{figure*}[]
\setlength{\abovecaptionskip}{-0.6cm}   %调整图片标题与图距离
\centering 
\includegraphics[width=\textwidth]{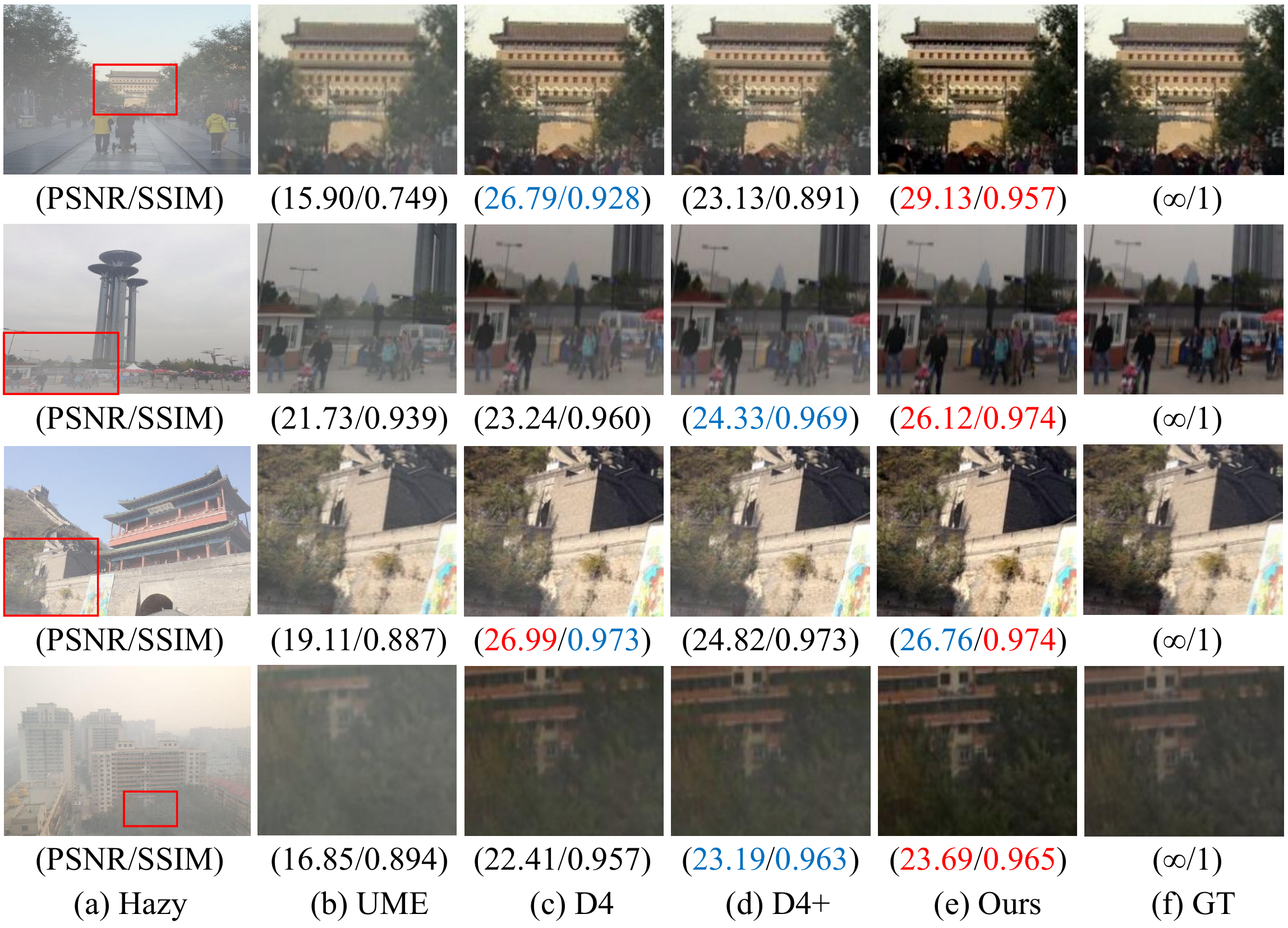} % 插入图片，假设图片文件名为 example-image.png
\caption{Visual comparison of haze removal on samples from the SOTS-outdoor dataset. It can be seen that, both in terms of visualization and metrics, our method performs the best. The \textcolor{red}{red} and \textcolor{blue}{blue} colors are used to indicate the $1^{st}$ and $2^{nd}$ ranks, respectively. } 
\label{fig:6} 
\end{figure*}

\begin{figure*}
\setlength{\abovecaptionskip}{-0.6cm}   %调整图片标题与图距离
\centering
\includegraphics[width=\textwidth]{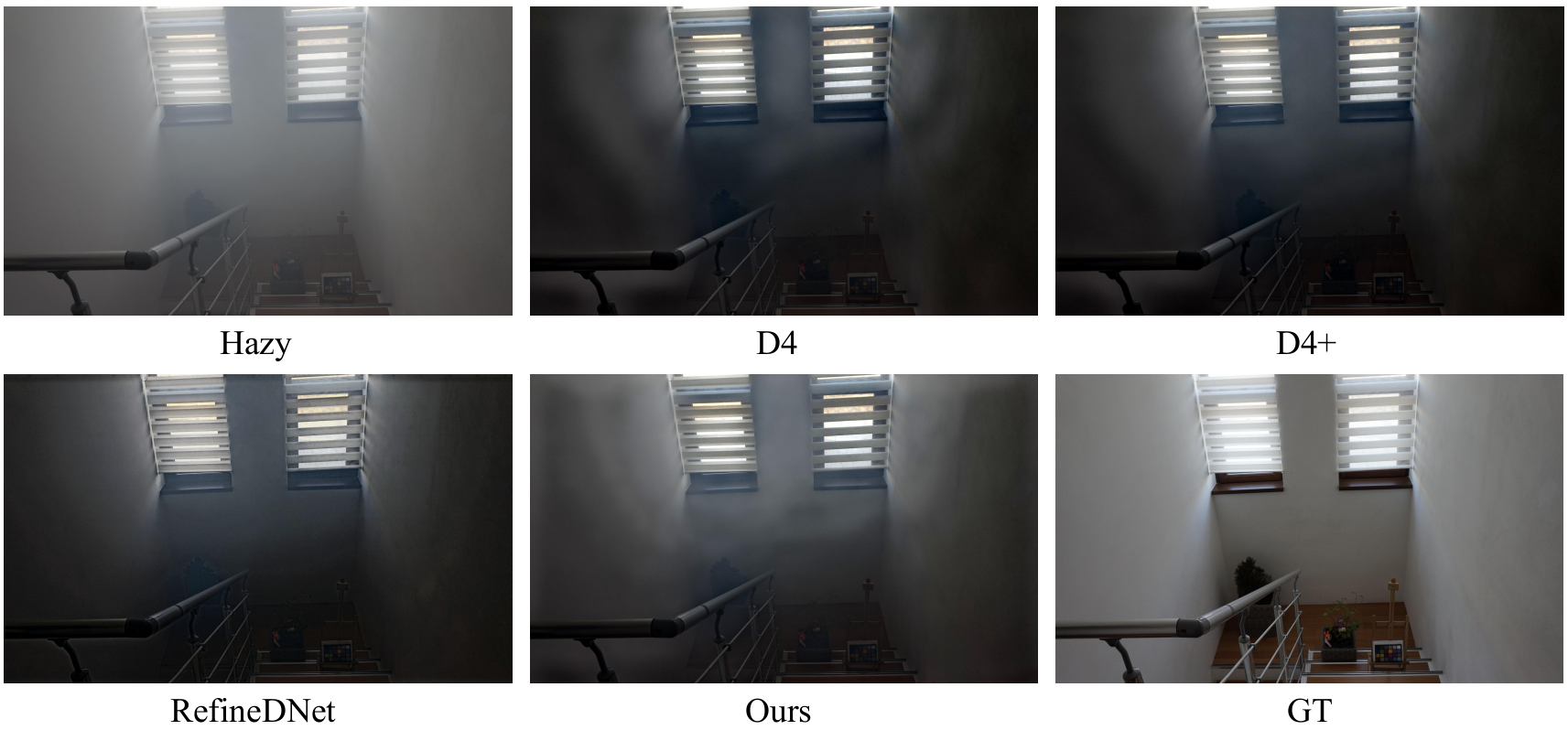}
\caption{Visual comparison of haze removal on samples from the I-HAZE dataset.}
\label{fig:7}
\end{figure*}

\begin{figure*}
\setlength{\abovecaptionskip}{-0.6cm}   %调整图片标题与图距离
\centering
\includegraphics[width=\textwidth]{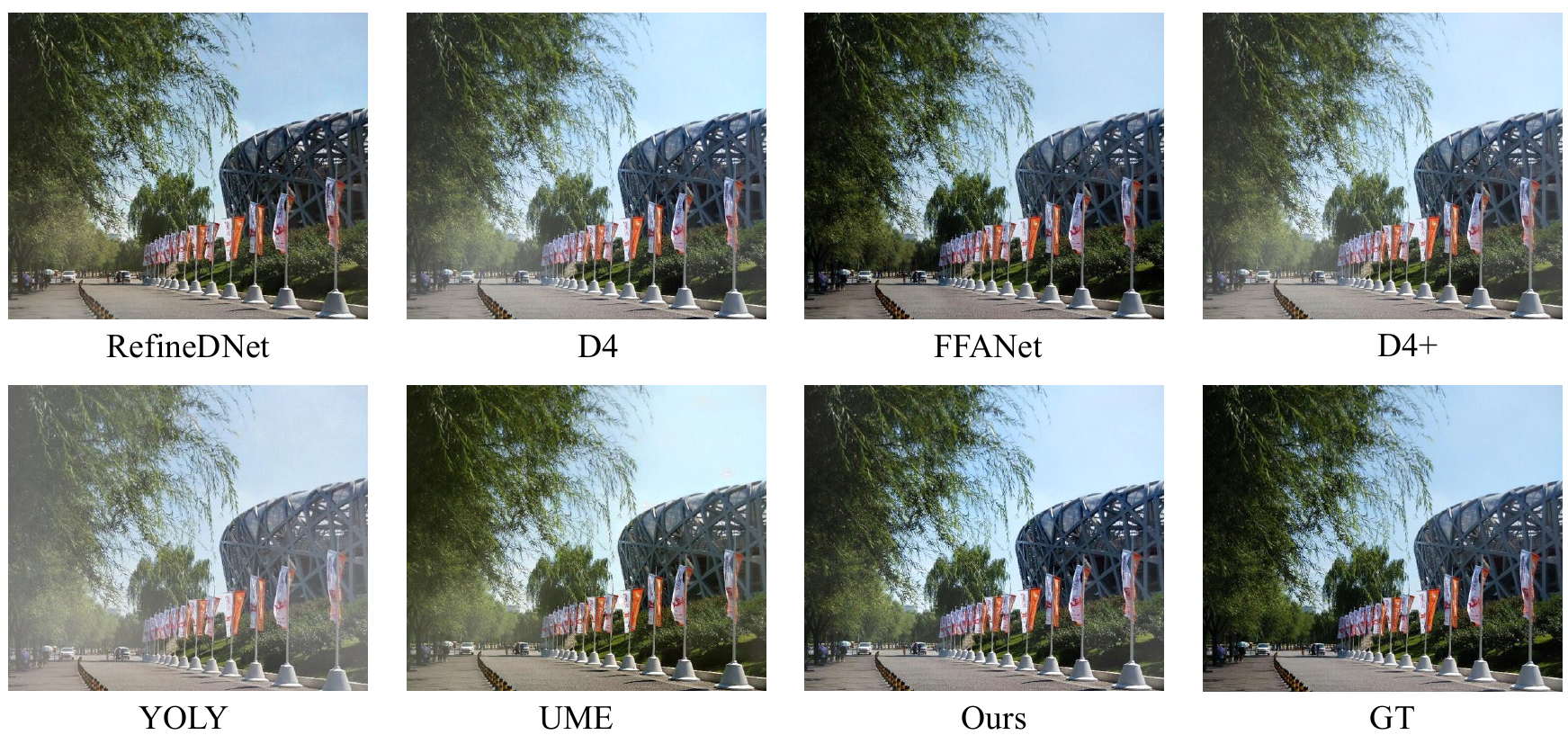}
\caption{Visual comparison of haze removal on samples from the HSTS dataset.}
\label{fig:8}
\end{figure*}

\section{Experiments}
\label{sec:section4}

In this section, we describe the experimental setup, dataset and evaluation metrics as outlined in Section~\ref{4.1}. Then, the quantitative and qualitative evaluation results of our method with more than 10 other excellent dehazing methods are shown and explained in Section~\ref{4.2}. In addition, in Section~\ref{4.3} we perform ablation experiments to validate the effectiveness of each module mentioned in the method. Finally, Section~\ref{4.4} discusses and analyzes the shortcomings of the model.

\subsection{Implementation Details}
\label{4.1}
\textbf{Experimental setup.} During the overall architecture training of UR2P-Dehaze, we used the Adam optimizer\cite{kingma2014adam}, where $\beta_1$ = 0.9, $\beta_2$ = 0.999, and the batch size was 8. The learning rate of the model is 0.0001. In addition, we randomly crop the image for 256×256 blocks for training. The entire network is executed on NVIDIA 3090 GPUs using the PyTorch framework.
\begin{figure*}[h]
\centering 
\setlength{\abovecaptionskip}{-0.6cm}   %调整图片标题与图距离
\includegraphics[width=\textwidth]{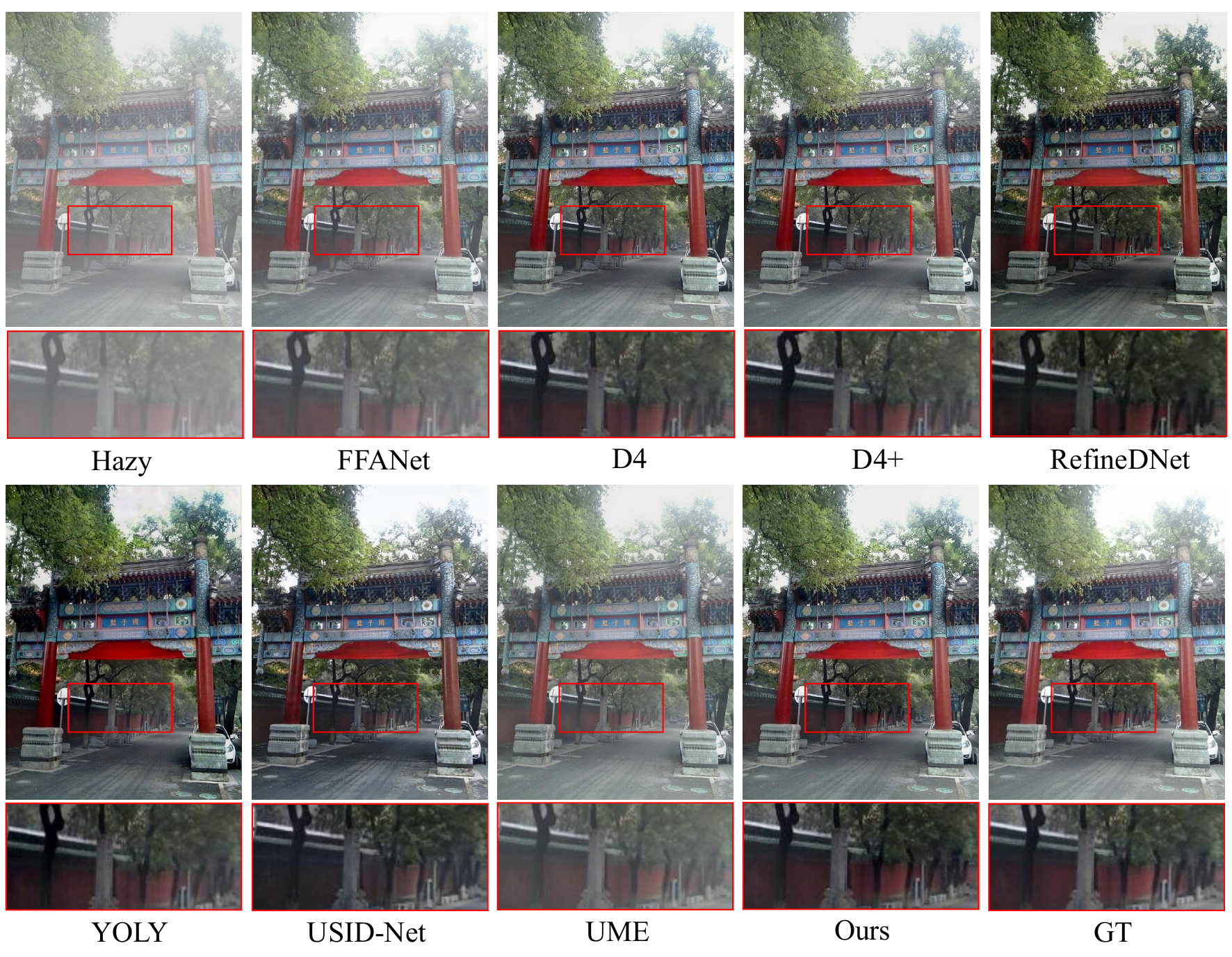} % 插入图片，假设图片文件名为 example-image.png
\caption{Visual comparison of haze removal on samples from the SOTS-outdoor dataset.} 

\label{fig:9} 
\end{figure*}

\textbf{The datasets.} To train the model, we used the RESIDE dataset\cite{li2018benchmarking}. RESIDE's indoor training dataset (ITS) contains 13,990 composite hazy images and 1,399 clear images, while the outdoor training dataset (OTS) contains 313,950 composite hazy images and 8,970 clear images. Our proposed approach is evaluated on both synthetic and real-world datasets. The datasets used for testing include a composite dataset of 500 indoor and 500 outdoor images, as well as an I-HAZE\cite{ancuti2018haze} dataset containing 35 images. The training datasets for our model are divided into indoor and outdoor, with ITS adopted for the indoor dataset. Due to resource limitations, 18,000 datasets are randomly selected from the original outdoor datasets OTS for training.
% \usepackage{multirow}
% \usepackage{multirow}

% \usepackage{multirow}
% \usepackage{multirow}

\textbf{Evaluation Metric.} To evaluate the performance of our methods, we selected three metrics for quantitative comparison: PSNR\cite{wang2004image}, SSIM\cite{wang2004image}, and CIEDE2000\cite{sharma2005ciede2000}. In addition, other perceptual indicators were compared, such as FID\cite{heusel2017gans} and LPIPS\cite{zhang2018unreasonable}. These metrics are commonly used to evaluate the effectiveness of dehazing networks. In addition, because some method codes are not open-source or the weights are not published, the method names in the comparison experiments are standardized by adding. '$ * $' in front of the method name indicates that the indicator data uses the results of the original article.

\subsection{Experimental Comparison}
\label{4.2}
\textbf{Qualitative analysis.} For outdoor datasets, this section qualitatively analyzes the performance difference between the proposed dehazing method and prior art through visual comparison. As shown in~\autoref{fig:6}, we selected several images with varying degrees of haze and scenes, and applied our method as well as several representative dehazing methods. In our experiments, the proposed dehazing method outperforms both traditional and existing techniques in several ways. Especially under high-haze conditions, our method not only successfully restores the contrast and saturation of the image but also avoids excessive enhancement, maintaining a natural feel. Additionally, as shown in the penultimate column of~\autoref{fig:6}, for detailed scenes, such as leaves and building edges, our method demonstrated superior detail retention, with sharper results and controlled halo effects compared to methods that could result in color distortion and edge blurring, thus preserving the overall natural feel and visual comfort of the image. These visual results demonstrate the effectiveness and superiority of our dehazing method. For the SOTS-indoor dataset, the visual results are shown in~\autoref{fig:indoor}. It can be observed that the images dehazed using the D4~\cite{yang2022self}, D4+~\cite{yang2024robust}, and RefineDNet~\cite{zhao2021refinednet} methods exhibit darker tones, while the dehazing effect of the LOLY~\cite{li2021you} method is relatively less effective. Notably, the dehazed images produced by our method are more closely aligned with the reference images. \par
Furthermore, as shown in~\autoref{fig:dect}, our method is compared with other SOTA methods that use the YOLO11n model for object detection. And as can also be seen in~\autoref{tab:dect-detail}, our method achieves impressive results in terms of confidence and detection accuracy.

\begin{table}[h]
\setlength{\abovecaptionskip}{0cm}
\setlength{\belowcaptionskip}{0.2cm}
\renewcommand{\arraystretch}{1.1}
\caption{Quantitative comparisons (Average PSNR/SSIM/LPIPS/FID/CIEDE2000) with dehazing approaches on the SOTS-indoor dataset. The \textcolor{red}{red} and \textcolor{blue}{blue} colors are used to indicate the $1^{st}$ and $2^{nd}$ ranks, respectively. “ $ * $ ” in front of the method name indicates that the indicator data uses the results of the original article.}
\scalebox{0.72}{

\begin{tabular}{lccccccccc}
\hline
                                            & Method     &  & Venue\&Year & \multicolumn{1}{l}{} & \multicolumn{5}{l}{SOTS-indoor}            \\ \cline{6-10} 
                                            &            &  &             & \multicolumn{1}{l}{} & PSNR$\uparrow$  & SSIM$\uparrow$  & LPIPS$\downarrow$ & FID$\downarrow$    & CIEDE2000$\downarrow$ \\ \hline
\multicolumn{1}{c}{\multirow{5}{*}{Paired}} & EPDN\cite{qu2019enhanced}       &  & CVPR 2019   &                      & 25.06 & 0.936 & 0.043 & 12.257 & 3.008     \\
\multicolumn{1}{c}{}                        & FFANet\cite{qin2020ffa}     &  & AAAI 2020   &                      & 36.36 & 0.990 & 0.009 & 4.869  & 0.274     \\
\multicolumn{1}{c}{}                        & SANet\cite{cui2023strip}      &  & IJCAI 2023  &                      & 40.34 & \textcolor{blue}{0.994} & 0.004 & 1.518  & \textcolor{blue}{0.165}     \\
\multicolumn{1}{c}{}                        & C2PNet\cite{zheng2023curricular}     &  & CVPR 2023   &                      & \textcolor{red}{42.37} & \textcolor{red}{0.995} & \textcolor{red}{0.002} & \textcolor{red}{1.160}  & \textcolor{red}{0.117}     \\
\multicolumn{1}{c}{}                        & DEANet\cite{chen2024dea}     &  & TIP 2024    &                      & \textcolor{blue}{41.22} & \textcolor{red}{0.995} & \textcolor{blue}{0.003} & \textcolor{blue}{1.279}  & 0.170     \\ \hline
\multirow{9}{*}{Unpaired}                 & RefineDNet\cite{zhao2021refinednet} &  & TIP 2021    &                      & 20.48 & 0.859 & 0.124 & 31.520 & 5.494     \\
                                            & YOLY\cite{li2021you}       &  & IJCV 2021   &                      & 15.27 & 0.534 & 0.189 & 36.628 & 8.054     \\
                                            & USID-NET\cite{li2022usid}    &  & TMM 2022     &                      & 17.42 & 0.813 & 0.248 & 55.598 & 14.471    \\
                                            & D4\cite{yang2022self}         &  & CVPR 2022   &                      & 25.42 & 0.930 & 0.045 & \textcolor{blue}{15.169} & 1.945     \\
                                            & *POGAN\cite{qiao2023learning}         &  & TCSVT 2023   &                      & 25.51 & 0.934 & - & - & -     \\
                                            & D4+\cite{yang2024robust}        &  & IJCV 2024   &                      & \textcolor{blue}{25.78} & \textcolor{blue}{0.934} & \textcolor{blue}{0.044} & 15.321 & \textcolor{blue}{1.894}     \\
                                            & *ADC-Net\cite{wei2024robust}        &  & TCSVT 2024   &                      & 25.52 & 0.935 & - & - & -     \\
                                            & *RPC-Dehaze\cite{lin2024toward}         &  &TCSVT 2024    &                      & 26.32 & 0.943 & - & - & -    \\
                                            & *ODCR\cite{wang2024odcr}         &  &CVPR 2024    &                      & 26.32 & 0.945 & - & - & -    \\

                                            & \textbf{Ours}       &  & /           &                      & \textcolor{red}{26.82} & \textcolor{red}{0.948} & \textcolor{red}{0.036} & \textcolor{red}{11.692} & \textcolor{red}{1.710}     \\ \hline
\end{tabular}
}
\label{tab:table1} % 添加标签
\end{table}

\begin{table}[h]
\setlength{\abovecaptionskip}{0cm}
\setlength{\belowcaptionskip}{0.2cm}
\renewcommand{\arraystretch}{1.1}
\caption{Quantitative comparisons (Average PSNR/SSIM/LPIPS/FID/CIEDE2000) with dehazing approaches on the SOTS-outdoor dataset. The \textcolor{red}{red} and \textcolor{blue}{blue} colors are used to indicate the $1^{st}$ and $2^{nd}$ ranks, respectively. “ $ * $ ” in front of the method name indicates that the indicator data uses the results of the original article.}

\scalebox{0.72}{
\begin{tabular}{lccccccccc}
\hline
                                            & Method     &  & Venue\&Year & \multicolumn{1}{l}{} & \multicolumn{5}{l}{SOTS-outdoor}            \\ \cline{6-10} 
                                            &            &  &             & \multicolumn{1}{l}{} & PSNR$\uparrow$   & SSIM$\uparrow$  & LPIPS$\downarrow$ & FID$\downarrow$    & CIEDE2000$\downarrow$ \\ \hline
\multicolumn{1}{c}{\multirow{5}{*}{Paired}} & EPDN\cite{qu2019enhanced}       &  & CVPR 2019   &                      & 20.30  & 0.887 & 0.104 & 16.165 & 8.491     \\
\multicolumn{1}{c}{}                        & FFANet\cite{qin2020ffa}     &  & AAAI 2020   &                      & 25.14  & 0.933 & \textcolor{blue}{0.036} & \textcolor{blue}{7.512}  & 4.114     \\
\multicolumn{1}{c}{}                        & SANet\cite{cui2023strip}      &  & IJCAI 2023  &                      & \textcolor{blue}{32.71}  & \textcolor{blue}{0.952} & 0.037 & 10.024 & \textcolor{red}{0.591}     \\
\multicolumn{1}{c}{}                        & C2PNet\cite{zheng2023curricular}     &  & CVPR 2023   &                      & 31.06  & 0.950 & \textcolor{blue}{0.036} & 9.261  & 1.064     \\
\multicolumn{1}{c}{}                        & DEANet\cite{chen2024dea}     &  & TIP 2024    &                      & \textcolor{red}{35.25} & \textcolor{red}{0.985} & \textcolor{red}{0.011} & \textcolor{red}{3.113}  & \textcolor{blue}{0.784}     \\ \hline
\multirow{10}{*}{Unpaired}                 & RefineDNet\cite{zhao2021refinednet} &  & TIP 2021    &                      & 20.87  & 0.906 & 0.088 & 18.323 & 6.643     \\
                                            & YOLY\cite{li2021you}       &  & IJCV 2021   &                      & 16.41  & 0.523 & 0.183 & 26.132 & 7.718     \\
                                            & USID-Net\cite{li2022usid}   &  & TMM 2022    &                      & 24.21  & 0.897 & 0.112 & 27.772 & 4.808     \\
                                            & D4\cite{yang2022self}         &  & CVPR 2022   &                      & 25.85  & 0.957 & 0.028 & 7.120  & 3.219     \\
                                            & *POGAN\cite{qiao2023learning}         &  & TCSVT 2023   &                      & 25.90  & 0.956 & - & -  & -     \\
                                            & *UCL\cite{wang2024ucl}         &  & TIP 2024   &                      & 25.21  & 0.927 & - & -  & 4.784     \\
                                            & D4+\cite{yang2024robust}        &  & IJCV 2024   &                      & 26.32  & 0.961 & \textcolor{blue}{0.025} & \textcolor{blue}{6.677}  & 3.062     \\
                                            & *RPC-Dehaze\cite{lin2024toward}        &  & TCSVT 2024   &                      & 26.41  &\textcolor{blue}{0.963}  & - & -  & -     \\
                                            & *ODCR\cite{wang2024odcr}         &  &CVPR 2024    &                      & 26.16 & 0.960 & - & - & -    \\
                                            & UME-NET\cite{sun2024unsupervised}    &  & PR 2024     &                      & \textcolor{blue}{27.26}  & 0.931 & 0.051 & 13.407 & \textcolor{blue}{2.720}     \\
                                            & \textbf{Ours}       &  & /           &                      & \textcolor{red}{27.53}  & \textcolor{red}{0.967} & \textcolor{red}{0.021} & \textcolor{red}{6.163}  & \textcolor{red}{2.457}     \\ \hline
\end{tabular}
}
\label{tab:table2} % 添加标签
\end{table}
\begin{table}[h]
\setlength{\abovecaptionskip}{0cm}
\setlength{\belowcaptionskip}{0.2cm}
\renewcommand{\arraystretch}{1.1}
\caption{Quantitative comparisons (Average PSNR/SSIM/LPIPS/FID/CIEDE2000) with dehazing approaches on the HSTS dataset. The \textcolor{red}{red} and \textcolor{blue}{blue} colors are used to indicate the $1^{st}$ and $2^{nd}$ ranks, respectively. “ $ * $ ” in front of the method name indicates that the indicator data uses the results of the original article.}
\scalebox{0.72}{
\begin{tabular}{lccccccccc}
\hline
                                            & \multicolumn{1}{l}{Method} &  & \multicolumn{1}{l}{Venue\&Year} & \multicolumn{1}{l}{} & \multicolumn{5}{l}{HSTS}                                                                                                \\ \cline{6-10} 
                                            &                            &  &                                 & \multicolumn{1}{l}{} & PSNR$\uparrow$   & \multicolumn{1}{l}{SSIM$\uparrow$} & \multicolumn{1}{l}{LPIPS$\downarrow$} & \multicolumn{1}{l}{FID$\downarrow$} & \multicolumn{1}{l}{CIEDE2000$\downarrow$} \\ \hline
\multicolumn{1}{c}{\multirow{5}{*}{Paired}} & EPDN\cite{qu2019enhanced}                       &  & CVPR 2019                       &                      &21.30        &0.867                          &0.112                           &49.976                         &6.900                               \\
\multicolumn{1}{c}{}                        & FFANet\cite{qin2020ffa}                     &  & AAAI 2020                       &                      & 30.128 & 0.939                    & 0.036                     & 26.532                  & 1.085                         \\ 
& C2PNet\cite{zheng2023curricular}                       &  & CVPR 2023                       &                      &31.12        &0.943                          &\textcolor{blue}{0.031}                           &\textcolor{red}{21.125}                         &0.828                               \\
& SANet\cite{cui2023strip}                       &  & IJCAI 2023                       &                      &\textcolor{red}{32.57}        &\textcolor{blue}{0.945}                          &0.033                           &22.756                         &\textcolor{blue}{0.568}                               \\
& DEANet\cite{chen2024dea}                       &  & TIP 2024                       &                      &\textcolor{blue}{32.55}        &\textcolor{red}{0.946}                          &\textcolor{red}{0.030}                           &\textcolor{blue}{21.967}                         &\textcolor{red}{0.453}                               \\
\hline
\multirow{10}{*}{Unpaired}                 & RefineDNet\cite{zhao2021refinednet}                 &  & TIP 2021                        &                      & 20.94 & 0.864                    & 0.116                     & 47.388                  & 5.148                         \\
                                            & YOLY\cite{li2021you}                       &  & IJCV 2021                       &                      & 16.74 & 0.483                    & 0.202                     & 64.770                  & 7.309                         \\
                                            & USID-NET\cite{sun2024unsupervised}                    &  & TMM 2022                         &                      &22.98        &0.754                          &0.137                           &69.554                         &3.923                               \\
                                            & D4\cite{yang2022self}                         &  & CVPR 2022                       &                      & 24.43 & 0.900                    & 0.056                     & 30.904                  & 4.276                         \\
                                            & *POGAN\cite{qiao2023learning}         &  & TCSVT 2023                       &                      & 24.11 & 0.917                    & -                     & -                  & -                         \\
                                            &*UCL\cite{wang2024ucl}                         &  & TIP 2024                       &                      & \textcolor{red}{26.87} & \textcolor{red}{0.933}                    &-                     &-                  &4.612                         \\
                                            & D4+\cite{yang2024robust}                        &  & IJCV 2024                       &                      & 24.43 & 0.903                    & 0.052                     & 31.423                  & 4.409                         \\
                                            & *RPC-Dehaze\cite{lin2024toward}                        &  & TCSVT 2024                       &                      & 25.76 & 0.958                    & -                     & -                  & -                         \\
                                            & UME-NET\cite{sun2024unsupervised}                    &  & PR 2024                         &                      &\textcolor{blue}{27.35}        &0.919                          &\textcolor{blue}{0.049}                           &\textcolor{blue}{29.977}                         &\textcolor{blue}{3.022}                               \\
                                            & \textbf{Ours}                       &  & /                               &                      & 26.71  & \textcolor{blue}{0.920}                    & \textcolor{red}{0.045}                     & \textcolor{red}{28.196}                  & \textcolor{red}{2.979}                         \\ \hline
\end{tabular}
}
\label{tab:table3} % 添加标签
\end{table}

\begin{table*}[h]
\setlength{\abovecaptionskip}{0cm}
\setlength{\belowcaptionskip}{0.2cm}
\renewcommand{\arraystretch}{1.1}
\caption{Quantitative comparisons (Average PSNR/SSIM/LPIPS/FID/CIEDE2000) with dehazing approaches on the I-HAZE dataset. The \textcolor{red}{red} and \textcolor{blue}{blue} colors are used to indicate the $1^{st}$ and $2^{nd}$ ranks, respectively. “ $ * $ ” in front of the method name indicates that the indicator data uses the results of the original article.}
\scalebox{0.76}{

\begin{tabular}{ccccccccc}
\hline
\multicolumn{1}{l}{Method} &  & \multicolumn{1}{l}{Venue\&Year} & \multicolumn{1}{l}{} & \multicolumn{5}{l}{I-HAZE}                                                                                             \\ \cline{5-9} 
                           &  &                                 & \multicolumn{1}{l}{} & PSNR$\uparrow$  & \multicolumn{1}{l}{SSIM$\uparrow$} & \multicolumn{1}{l}{LPIPS$\downarrow$} & \multicolumn{1}{c}{FID$\uparrow$} & \multicolumn{1}{l}{CIEDE2000$\downarrow$} \\ \hline
RefineDNet\cite{zhao2021refinednet}                 &  & TIP 2021                        &                      & 15.50 & 0.710                    & 0.407                     & 142.328                 & 10.395                         \\
D4\cite{yang2022self}                          &  & CVPR 2022                      &                      & \textcolor{blue}{15.71} & \textcolor{blue}{0.753}                    & 0.364                     & \textcolor{blue}{130.458}                 &9.910                               \\
*Cycle-SNSPGAN\cite{wang2022cycle}                          &  & TITS 2022                      &                      & 15.36 & 0.740                    & -                     & -                 &-                               \\
*IC-Dehazing\cite{gui2023illumination}                          &  & TMM 2023                      &                      & 15.70 & 0.726                    & -                     & -                 &-                               \\
D4+\cite{yang2024robust}                        &  & IJCV 2024                       &                      & 15.61 & 0.749                    & \textcolor{blue}{0.357}                     & \textcolor{red}{128.765}                 &\textcolor{blue}{9.893}                               \\
*RPC-Dehaze\cite{lin2024toward}                        &  & TCSVT 2024                       &                      & 16.26 & 0.783                    & -                     & -                 &-                               \\
\textbf{Ours}                       &  & /                               &                      & \textcolor{red}{16.36} & \textcolor{red}{0.770}                    & \textcolor{red}{0.345}                     & 161.615                 &\textcolor{red}{8.158}                               \\ \hline
\end{tabular}
}
\label{tab:table4} % 添加标签
\end{table*}

\textbf{Quantitative analysis.} In terms of quantitative analysis, our haze removal method has demonstrated excellent performance on multiple evaluation indicators. Among them, based on the test results of the SOTS-outdoor dataset, as shown in ~\autoref{tab:table2}, the contrasting methods include supervised methods such as EPDN\cite{qu2019enhanced}, FFANet\cite{qin2020ffa}, SANet\cite{cui2023strip}, C2PNet\cite{zheng2023curricular} and DEANet\cite{chen2024dea}. Unsupervised methods such as   YOLY\cite{li2021you}, RefineDNet\cite{zhao2021refinednet}, RPC-Dehaze\cite{lin2024toward}, D4+\cite{yang2024robust}, UCL\cite{ wang2024ucl}, POGAN\cite{qiao2023learning}, USID-Net\cite{li2022usid}, ODCR\cite{wang2024odcr}, D4\cite{yang2022self} and UME-Net\cite{ sun2024unsupervised}. As shown in~\autoref{fig:6}--\ref{fig:9}, through visual comparison, our method effectively restores the natural colors and details of the image while significantly reducing artifacts and distortions that may arise during the de-ghosting process. According to the test metrics, our method shows a significant improvement in image quality restoration compared to the baseline method, (a) For the sots-outdoor dataset, the PSNR improves by 1.68 dB, and the SSIM improves by 0.036. (b) For the sots-indoor dataset, the PSNR improves by 1.4 dB, and the SSIM improves by 0.018. (c) For the HSTS dataset, PSNR improves by 2.28 dB and SSIM improves by 0.02. (d) For the I-HAZE dataset, PSNR improves by 0.65 dB and SSIM improves by 0.017. More experimental data are shown in~\autoref{tab:table1}--\ref{tab:table4}. In addition, there is a significant decrease in LPIPS, FID and CIEDE2000. There is also a significant improvement compared to the SOTA method.

\begin{figure*}[h]

\centering 
\includegraphics[width=0.9\textwidth]{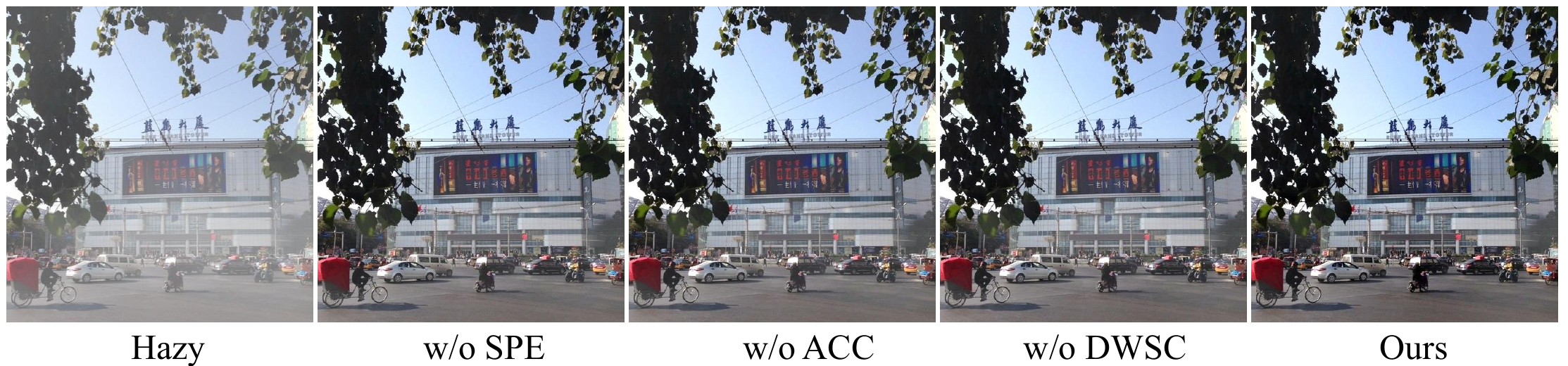} % 插入图片，假设图片文件名为 example-image.png
\caption{Visual comparisons of the ablation studies.} 
\label{fig:xiaorong} 
\end{figure*}
\subsection{Ablation Experiment}
\label{4.3}
To demonstrate the effectiveness of our proposed UR2P-Dehaze method, we perform ablation studies on SPE, DWSC and ACC modules in each of the four public datasets.

\textbf{The Effectiveness of Shared Prior Estimator.}
To verify the beneficial impact of shared prior Estimator 
 (SPE), we conducted comparative experiments, as shown in~\autoref{tab:i-haze_hsts-ablation}. The results of the study indicate that using only a single deep prior knowledge is limited in terms of performance. In contrast, the model combining illumination and reflectance outperforms the model using either prior information alone. This is due to the limited parameter information that can be learned from hazy image, which makes it difficult to enhance recovery. In contrast, the estimation provides additional guidance to the network, achieving the best final performance.

\begin{figure*}[h]
\setlength{\abovecaptionskip}{-0.6cm}   %调整图片标题与图距离
\centering 
\includegraphics[width=\textwidth]{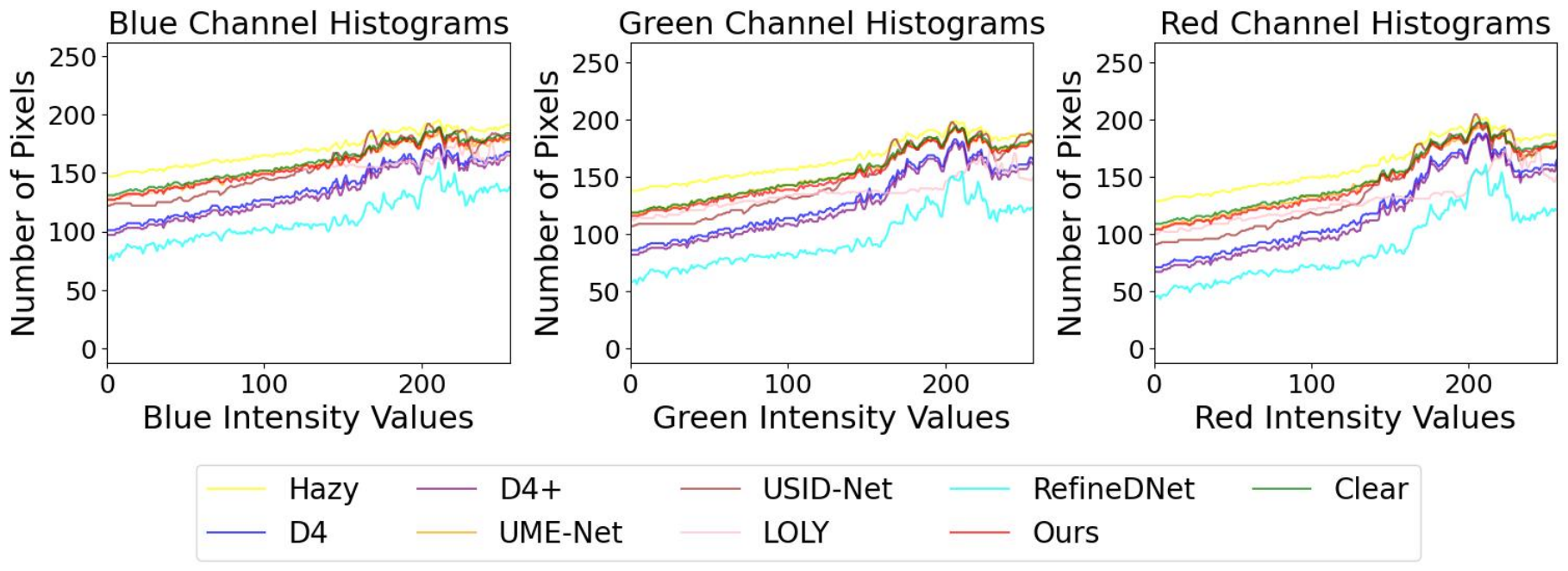} % 插入图片，假设图片文件名为 example-image.png
\caption{The impact of different dehazing methods (such as D4, D4+, UME-Net, etc.) on the pixel intensity distribution of three color channels (Blue, Green, and Red). The histograms provide a visual comparison of each method's performance in restoring the distribution closer to that of the clear reference image.} 
\label{fig:fig11} 
\end{figure*}

\textbf{The Effectiveness of Adaptive Color Corrector.}
The results of an ablation study with an adaptive color Corrector (ACC) are reported, as shown in ~\autoref{tab:sots-ablation}. In this study, we initially excluded the shared prior estimator (SPE) and dynamic wavelet separable convolution (DWSC), forming a baseline model by adding an ACC. The experimental results show that the adaptive color corrector brings a significant performance improvement across four different datasets. This result confirms the importance of the adaptive color corrector in improving image quality. As shown in~\autoref{fig:fig11}, the green line represents the clear image, while the image after defogging by our method is the red line. It can be noticed that the two almost overlap on the RGB three channels. That is to say, it shows that the model using an adaptive color corrector can better restore the color and make the image closer to the real scene, thus verifying its effectiveness.
\begin{table*}[h]
\setlength{\abovecaptionskip}{0cm}
\setlength{\belowcaptionskip}{0.2cm}
\renewcommand{\arraystretch}{1.1}
\caption{Ablation study of our proposed module on the I-HAZE and HSTS dataset. The best performance will be highlighted in \textbf{bold}.}
\scalebox{0.32}{
\fontsize{27}{25}\selectfont % 调整字体大小
\begin{tabular}{lccccccccc}
\hline
Method               & \multicolumn{4}{l}{I-HAZE}        & \multicolumn{1}{l}{} & \multicolumn{4}{l}{HSTS}          \\ \cline{2-5} \cline{7-10} 
                     & PSNR$\uparrow$  & SSIM$\uparrow$  & LPIPS$\downarrow$ & CIEDE2000$\downarrow$ & \multicolumn{1}{l}{} & PSNR$\uparrow$  & SSIM$\uparrow$  & LPIPS$\downarrow$ & CIEDE2000$\downarrow$ \\ \cline{1-5} \cline{7-10} 
Baseline             & 15.71 & 0.753 & 0.407 & 10.395    &                      & 24.43 & 0.900 & 0.056 & 4.276      \\
w SPE          &15.92       &0.751       &0.350       &8.656           &                      &24.74       &0.901       &0.053       &4.444           \\
w DWSC        & 15.82 & 0.756 & 0.351 & 8.480     &                      & 24.35 & 0.898 & 0.053 & 4.690     \\
w ACC         & 16.03 & 0.763 & 0.348 & 8.328     &                      & 25.43 & 0.906 & 0.050 & 4.001     \\
w/o ACC         &15.93     &0.754     &0.335     &8.273         &                      &24.93      &0.899      &0.055     &4.359          \\
w/o SPE         &12.83      &0.608      &0.441     &15.484      &                      &25.41    &0.910     &0.050      &3.599      \\
w/o DWSC      & 16.10 & 0.757 & 0.349 & \textbf{8.098}    &                      & 24.02 & 0.894 & 0.055 & 4.273     \\
\textbf{UR2P-Dehaze} & \textbf{16.36} & \textbf{0.770} & \textbf{0.345} & 8.158     &                      & \textbf{26.71} & \textbf{0.920} & \textbf{0.045} & \textbf{2.979}     \\ \hline
\end{tabular}
}

\label{tab:i-haze_hsts-ablation} % 添加标签
\end{table*}

\begin{table*}[h]
\setlength{\abovecaptionskip}{0cm}
\setlength{\belowcaptionskip}{0.2cm}
\renewcommand{\arraystretch}{1.1}
\caption{Ablation study of our proposed module on the SOTS-indoor and SOTS-outdoor dataset. The best performance will be highlighted in \textbf{bold}.}
\scalebox{0.32}{
\fontsize{27}{25}\selectfont % 调整字体大小
\begin{tabular}{lccccccccc}
\hline
Method               & \multicolumn{4}{l}{SOTS-indoor}   & \multicolumn{1}{l}{} & \multicolumn{4}{l}{SOTS-outdoor}  \\ \cline{2-5} \cline{7-10} 
                     & PSNR$\uparrow$  & SSIM$\uparrow$  & LPIPS$\downarrow$ & CIEDE2000$\downarrow$ & \multicolumn{1}{l}{} & PSNR$\uparrow$  & SSIM$\uparrow$  & LPIPS$\downarrow$ & CIEDE2000$\downarrow$ \\ \cline{1-5} \cline{7-10} 
Baseline             & 25.42 & 0.930 & 0.045 & 1.945     &                      & 25.85 & 0.957 & 0.028 & 3.219     \\
w SPE          &25.95       &0.940       &0.038       &1.902           &                      &26.49       &0.959       &0.024       &3.018           \\
w DWSC        & 25.95 & 0.938 & 0.039 & 2.080     &                      & 26.53 & 0.959 & 0.024 & 3.151     \\
w ACC         & 25.75 & 0.940 & 0.038 & 2.068     &                      & 26.24 & 0.957 & 0.025 & 3.061     \\
w/o ACC         &25.17     &0.929     &0.040     &2.590         &                      &26.19      &0.955      &0.026     &3.226          \\
w/o SPE         &25.97      &0.938      &0.038     &1.985      &                      &25.91    &0.955     &0.028      &3.166      \\
w/o DWSC      & 26.06 & 0.941 & 0.039 & 1.962     &                      & 26.40 & 0.958 & 0.025 & 3.087     \\
\textbf{UR2P-Dehaze}  & \textbf{26.82} & \textbf{0.948} & \textbf{0.036} & \textbf{1.710}     &                      & \textbf{27.53} & \textbf{0.967} & \textbf{0.021} & \textbf{2.457}     \\ \hline
\end{tabular}
}
\label{tab:sots-ablation} % 添加标签
\end{table*}
\begin{figure*}[h]
\setlength{\abovecaptionskip}{-0.6cm}   %调整图片标题与图距离
\centering 
\includegraphics[width=\textwidth]{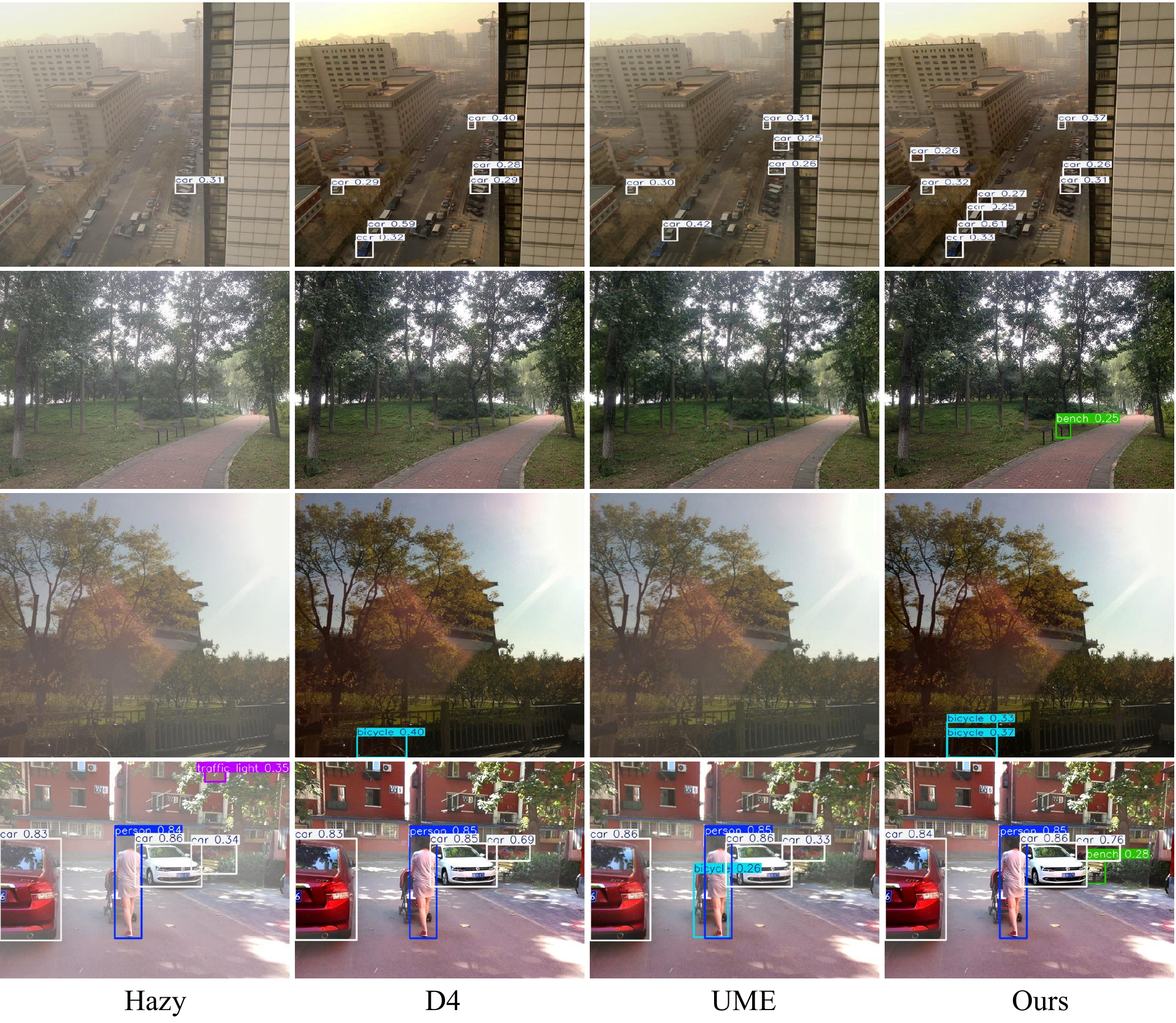} % 插入图片，假设图片文件名为 example-image.png
\caption{Comparison with other SOTA methods on object detection. In the first three images, many targets failed to be detected successfully, while the last image exhibited false detections.} 
\label{fig:dect} 
\end{figure*}
\begin{table*}[h]
\setlength{\abovecaptionskip}{0cm}
\setlength{\belowcaptionskip}{0.2cm}
\renewcommand{\arraystretch}{1}
\caption{Differences between various methods in the RGB channels compared to the original image. The \textcolor{red}{red} and \textcolor{blue}{blue} colors denote the $1^{st}$ and $2^{nd}$ ranks, respectively, based on the results from the SOTS-indoor dataset.} % 添加标题
\begin{center} % 使用center环境
\resizebox{0.4\textwidth}{!}{\begin{tabular}{cccc}
\hline
\textbf{Method}     & \textbf{R}     & \textbf{G}     & \textbf{B}     \\ \hline
Hazy       & 62.70 & 60.63 & 52.41 \\
RefineDNet\cite{zhao2021refinednet} & 21.46 & 20.25 & 18.27 \\
YOLY\cite{li2021you}       & 35.19 & 34.70 & 31.84 \\
D4\cite{yang2022self}         & 11.12 & 10.76 & 9.31  \\
D4+\cite{yang2024robust}        & \textcolor{blue}{10.57} & \textcolor{blue}{10.22} & \textcolor{blue}{8.86}  \\
\textbf{Ours}       & \textcolor{red}{9.13}  & \textcolor{red}{8.86}  & \textcolor{red}{7.70}  \\ \hline
\end{tabular}}
\end{center}
\label{tab:RGB_diff} % 添加标签
\end{table*}
\begin{table*}[h]
\centering
\setlength{\abovecaptionskip}{0cm}
\setlength{\belowcaptionskip}{0.2cm}
\renewcommand{\arraystretch}{1.1}
\caption{A comparison of object detection is conducted with hazy images, baseline methods, SOTA methods, and our method after dehazing.}
\scalebox{0.82}{
\begin{tabular}{c|lllllc}
\hline
Method                           & Category & Hazy                                                                       & D4                                                      & UME                                                                 & Ours                                                               & Analysis      \\ \hline
\multirow{6}{*}{Before Dehazing} & Row 1    & 1 car                                                                      & 6 car                                                   & 5 car                                                               & 9 car                                                              & More accurate \\ \cline{2-7} 
                                 & Row 2    & 0 bench                                                                    & 0 bench                                                 & 0 bench                                                             & 1 bench                                                            & Detectable    \\ \cline{2-7} 
                                 & Row 3    & 0 bicycle                                                                  & 1 bicycle                                               & 0 bicycle                                                           & 2 bicycle                                                          & Detectable    \\ \cline{2-7} 
                                 & Row 4    & \begin{tabular}[c]{@{}l@{}}3 car\\ 1 person\\ 1 traffic light\end{tabular} & \begin{tabular}[c]{@{}l@{}}3 car\\ 1 person\end{tabular} & \begin{tabular}[c]{@{}l@{}}3 car\\ 1 person\\ 1 bicycle\end{tabular} & \begin{tabular}[c]{@{}l@{}}3 car\\ 1 person\\ 1 bench\end{tabular} & More accurate \\ \hline
\end{tabular}
\label{tab:dect-detail}
}
\end{table*}
\begin{figure*}[h]
\setlength{\abovecaptionskip}{-0.6cm}   %调整图片标题与图距离
\centering 
\includegraphics[width=\textwidth]{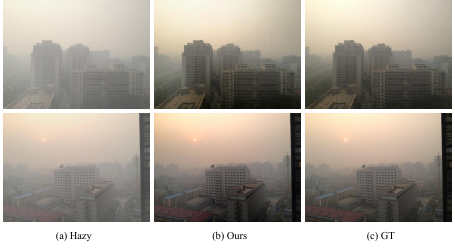} % 插入图片，假设图片文件名为 example-image.png
\caption{Extreme weather conditions cases. (a) The hazy image, (b) The dehazed image by our method, and (c) The clear image.} 
% \vspace{-10pt}
\label{fig:heavy_hazy} 
\end{figure*}

\textbf{Analysis of differences.} In the comparative analysis of the processed images of different haze removal methods with the RGB three-channel of the reference image, we observed significant performance differences, which are mainly due to the scattering of light by the haze resulting in color distortion and contrast degradation. In contrast, although the D4\cite{yang2022self} and D4+\cite{yang2024robust} methods have made some progress in removing the haze, there is still room for improvement in color restoration as shown in~\autoref{tab:RGB_diff}, as evidenced by their relatively high values in the RGB channel.

\textbf{The Effectiveness of Dynamic Wavelet Separable Convolution.}
To demonstrate the multi-scale feature extraction capability of our designed Dynamic Wavelet Separable Convolution (DWSC), we evaluated our UR2P-Dehaze network on four widely used synthetic datasets. Compared to other methods, ours shows a significant advantage in color restoration, further proving the effectiveness of DWSC. As illustrated in~\autoref{fig:xiaorong}, DWSC not only enhances the overall image quality but also significantly improves visual perception. Notably, DWSC leverages the properties of wavelet transforms to perform efficient convolution operations across different frequency bands. This enables it to capture local features in the image effectively while maintaining computational efficiency.\par
\subsection{Discussion}
\label{4.4}
We propose that the UR2P-Dehaze method effectively overcomes the limitations of existing methods in terms of adaptation and generalization. Experimental results show that UR2P-Dehaze performs well on several public datasets. However, as shown in~\autoref{fig:heavy_hazy}, the shortcomings are mainly in: (a) In extreme weather, the density of haze in the image is too thick to be handled well by any of the existing methods, which is mainly due to the limited information provided by the image. (b) The existing training data is of low quality, which may lead to unstable training. We hope that future research can further optimize its robustness and adaptability to improve performance in more complex environments.

\section{Conclusion}
\label{sec:section5}
We propose UR2P-Dehaze, a novel adaptive dehazing framework for learning unpaired rich a priori information. First, the framework is guided by Retinex theory in training and uses iterative optimization to ensure that the network output matches the clear image content of the input hazy image. Meanwhile, a Dynamic Wavelet Separable Convolution is designed in the enhanced network, which expands the receptive field in the wavelet domain and optimizes feature extraction at different scales. Finally, in the process of image reconstruction, to recover the color information more accurately, we designed an Adaptive Color Corrector, which enhances the realism of the recovered image. In conclusion, compared with traditional dehazing methods, UR2P-Dehaze does not rely on artificial features or physical models; it automatically extracts a priori information from hazy image and restores visual effects realistically, offering efficient and easy-to-deploy unsupervised dehazing. This approach provides new perspectives for low-level vision tasks.

\bibliographystyle{elsarticle-num} 
\bibliography{cas-refs.bib}

%\begin{thebibliography}{10}



% %% For numbered reference style

%   \textit{\LaTeX: a document preparation system},
%   Addison Wesley, Massachusetts,

% \end{thebibliography}

% \bibliographystyle{plain}
% \bibliography{cas-refs.bib} 

% \end{thebibliography}

\end{document}